\documentclass[lettersize,journal]{IEEEtran}
\usepackage{amsmath,amsfonts}
\usepackage{algorithmic}
\usepackage{algorithm}
\usepackage{array}
\usepackage[caption=false,font=normalsize,labelfont=sf,textfont=sf]{subfig}
\usepackage{textcomp}
\usepackage{stfloats}
\usepackage{url}
\usepackage{verbatim}
\usepackage{graphicx}
\usepackage{cite}
\usepackage{multirow}
\usepackage{makecell}
\usepackage{color}
\usepackage{booktabs}
\usepackage{enumitem}
\usepackage{float}
\usepackage{needspace}
\usepackage{ulem}
\usepackage[colorlinks=true]{hyperref}
\usepackage[capitalize]{cleveref}
\crefname{section}{Sec.}{Secs.}
\Crefname{section}{Section}{Sections}
\Crefname{table}{Table}{Tables}
\crefname{table}{Tab.}{Tabs.}
\def\name{GenMesh}
\def\eg{\textit{e}.\textit{g}.}
\def\ie{\textit{i}.\textit{e}.}
\def\etal{\textit{et al}.}
\newcommand{\revise}{}

\begin{document}

\title{\revise{Single-view 3D Mesh Reconstruction for Seen and Unseen Categories}}

%         <-this % stops a space
\author{Xianghui Yang, Guosheng Lin, Luping Zhou}
\markboth{Journal of \LaTeX\ Class Files,~Vol.~14, No.~8, August~2021}%
{Shell \MakeLowercase{\textit{et al.}}: A Sample Article Using IEEEtran.cls for IEEE Journals}


\maketitle

\begin{abstract}
Single-view 3D object reconstruction is a fundamental and challenging computer vision task that aims at recovering 3D shapes from single-view RGB images. Most existing deep learning based reconstruction methods are trained and evaluated on the same categories, and they cannot work well when handling objects from novel categories that are not seen during training. Focusing on this issue, this paper tackles \revise{Single-view 3D Mesh Reconstruction}, to study the model generalization on unseen categories and encourage models to reconstruct objects literally. Specifically, we propose an end-to-end two-stage network, \name, to break the category boundaries in reconstruction. Firstly, we factorize the complicated image-to-mesh mapping into two simpler mappings, \ie, image-to-point mapping and point-to-mesh mapping, while the latter is mainly a geometric problem and less dependent on object categories. Secondly, we devise a local feature sampling strategy in 2D and 3D feature spaces to capture the local geometry shared across objects to enhance model generalization. Thirdly, apart from the traditional point-to-point supervision, we introduce a multi-view silhouette loss to supervise the surface generation process, which provides additional regularization and further relieves the overfitting problem. The experimental results show that our method significantly outperforms the existing works on the ShapeNet and Pix3D under different scenarios and various metrics, especially for novel objects. The project link is \url{https://github.com/Wi-sc/GenMesh}.
\end{abstract}


\section{Introduction}
\label{sec:intro}
 \IEEEPARstart{H}{umans} are able to imagine a rough 3D shape from a given RGB image even though the particular object was not seen before. Modeling such a reconstruction process is an interesting research topic in computer vision, known as the single-view 3D reconstruction. This task is highly ill-posed because it is unable to know the object portion unseen from the camera perspective. Humans can accomplish the task because we store plenty of objects in mind, which enables us to establish the correlation between 2D images and 3D shapes. The correlation can further be applied to novel images carrying familiar objects. Based on this observation, computer vision researchers turn to deep learning techniques to mimic the 3D reconstruction process of humans. Recently, deep neural networks have achieved remarkable progress on a variety of 3D object representations, \eg, voxels~\cite{3DR2N2,occnet,richter2018matryoshka,CoReNet,Probabilistic_Latent_Space}, point clouds~\cite{CD_EMD_LOSS,han2020drwr,chao2021unsupervised}, and implicit functions~\cite{ImplicitFields,PIFu,PIFuHD}, which are friendly to current network architectures. However, compared with the above representations, the mesh based representation has not yet been fully explored even though it is more efficient to store geometry information and more widely used in industries. 
 
\begin{figure}[t]
% \vspace{-40pt}
\begin{center}
  \includegraphics[width=1.\linewidth]{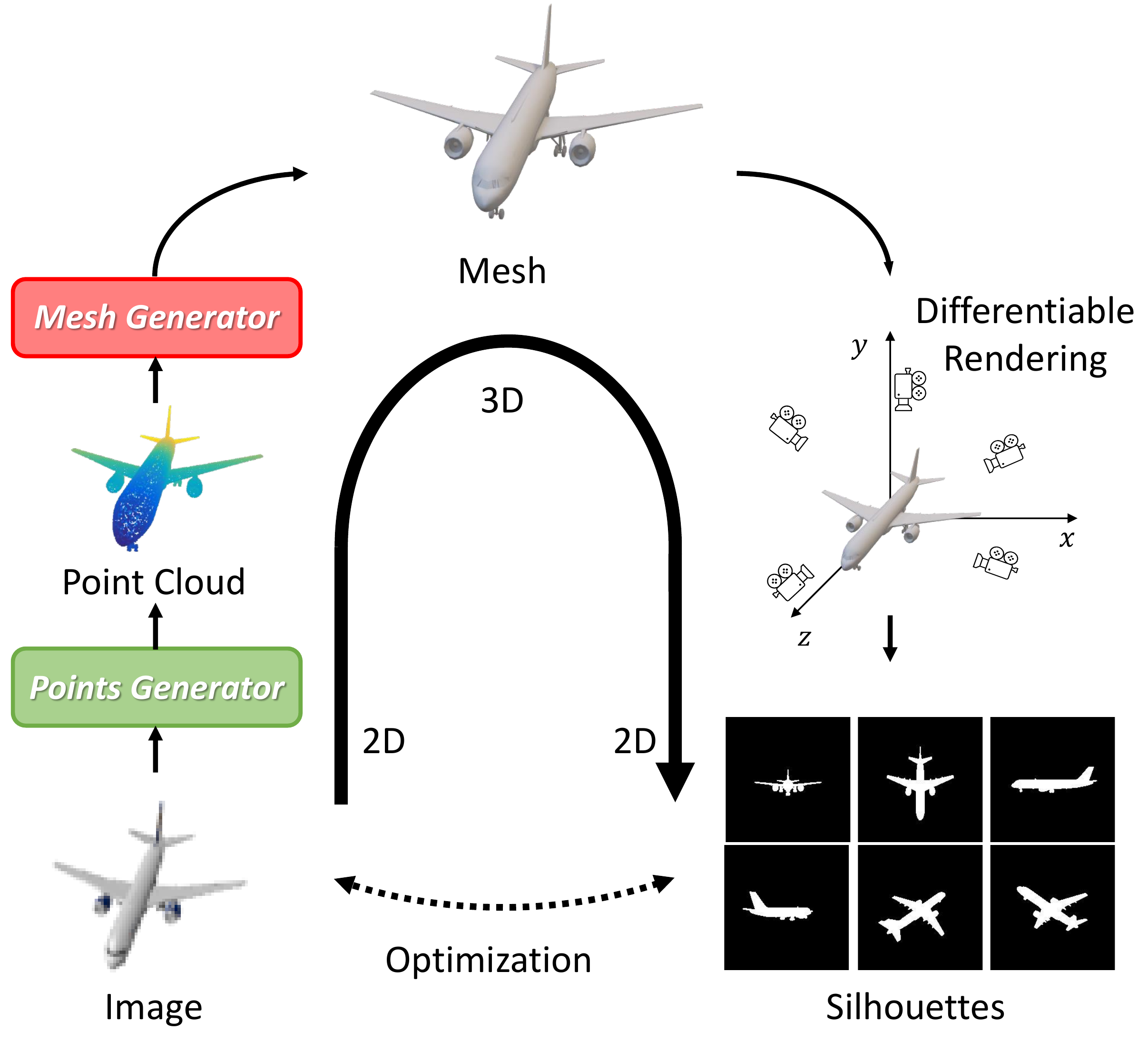}
%   \vspace{-10pt}
  \caption{Overview of the proposed \name\ framework for single-view 3D mesh reconstruction. It consists of two stages jointly trained: i) point cloud generation from the single input image, and ii) mesh generation from the intermediate point cloud.}
  \end{center}
\label{fig:head}
% \vspace{-20pt}
\end{figure}
 
Existing single-view 3D mesh reconstruction works mostly adopt the encoder-decoder architectures, where the encoder extracts perceptual features from the input images while the decoder deforms a template (2D square~\cite{atlasnet} or sphere~\cite{pixel2mesh,tmn}) to warp the target 3D shape. %Typically, the decoder predicts the offset of each vertex and preserves connections (edges) of the template(s). 
These mesh reconstruction networks are normally trained and evaluated on the same categories with encouraging performance. However, when generalizing them to novel categories, we observe a substantial performance decline as these training protocols suffer from the overfitting problem on the seen classes.
 %they are all limited by the categories inevitably, \ie, bad generalization on objects from novel categories. 
 %This training protocol also leads to overfitting on seen classes. More importantly, 
 The class-specific models are difficult to deploy in practice since it is inherently infeasible to collect a training set covering all types of object shapes around the world. Moreover, handling objects from various categories will also cause performance degradation, even when these categories are included in the training set. Thus, an important question was raised: what do single-view 3D reconstruction networks indeed learn? Tatarchenko~\etal~\cite{whatdolearn} argue that the existing models actually function by recognition rather than reconstruction, which is supported by an observation that a benchmark work~\cite{atlasnet} performs even worse than the simple nearest neighbor searching or classification baselines, both quantitatively and qualitatively.
%although convolutional neural networks for single-view 3D reconstruction have shown impressive performance, all of them actually conduct recognition instead of reconstruction. 

To improve model generalization in single-view 3D mesh reconstruction, this paper aims to tackle the ~\revise{single-view 3D mesh reconstruction on seen and unseen categories}. We develop a new framework, \textbf{\name}, to encourage the model to reconstruct 3D shapes literally instead of retrieving objects from memory. Under this setting, the model is trained on seen classes, \ie, \textbf{base classes}, and tested on unseen classes, \ie, \textbf{novel classes}. \name~comprises a two-stage pipeline to generate the triangular mesh from a single-view RGB image, with three strategies to break the boundaries of object categories for better model generalization.

\underline{\textbf{First}}, we disentangle the complicated reconstruction process, \ie, image-to-mesh mapping, into two simpler mappings, \ie, image-to-point mapping and point-to-mesh mapping. This operation decreases the complexity of the learning task and makes the training process easier. Moreover, the point-to-mesh mapping is more of a geometric problem than a recognition problem and has less dependency on object categories. Evidently, network models for 3D point meshing (point-to-mesh generation) showed good performance on novel classes~\cite{atlasnet}. Specifically, this strategy is implemented in a two-stage framework trained in an end-to-end manner. At the first stage, a point cloud generator estimates the point clouds of the target objects guided by image features. At the second stage, a mesh generator deforms triangular templates into target meshes by vertex shifts towards the intermediate point cloud, while preserving raw connections during deformation. 
\underline{\textbf{Second}}, we explore the feature representations, \ie, local features ~\cite{pixel2mesh,MeshRCNN} and global features ~\cite{3DR2N2,atlasnet,tmn,genre,GSIR,sdfnet,few_shot_voxel_2019,few_shot_voxel_2020}, for the generalization, proving the superiority of feature sampling strategy instead of the widely-used global features, and extend the 2D local feature sampling into 3D to prevent the model from overfitting. We argue that global features describing overall shapes easily overfit training object categories and thus cannot generalize well to novel categories. In contrast, local features focus on local geometry shared across categories (seen and unseen) and prevent recognition by the limited receptive field. Although local features have been sporadically seen in single-view 3D reconstruction~\cite{pixel2mesh, MeshRCNN}, they were mainly proposed to improve reconstruction quality, rather than being explicitly linked to model generalization. Evidently, most reconstruction methods studying model generalization~\cite{sdfnet,genre,GSIR} still adopt global features. This paper would like to emphasize that local features could not only improve the details of reconstruction but also matter to model generalization, especially for novel unseen object categories, as carefully evaluated in our work. It attempts to partially answer the question in~\cite{whatdolearn}: how to encourage models to really learn reconstruction instead of retrieval. Specifically, the 2D Feature Sampler obtains local features from the extracted 2D feature maps by 3D-to-2D projection according to camera intrinsic matrix and pose, while the 3D Feature Sampler extracts local features from the encoded 3D feature groups of the generated point cloud by nearest neighbor searching. 
\underline{\textbf{Third}}, we design a multi-view silhouette loss in 2D space to inject the face quality supervision, which serves as an additional regularization for reconstruction. It is noted that the widely used point-to-point distance, Chamfer Distance, is not sufficient to supervise the edges and faces generation due to the lack of a direct measure for evaluation. Thus, we introduce additional supervision based on the consistency of the 2D projections of the predicted and the ground-truth meshes from different camera views. It is noteworthy that although 2D local features and silhouette loss have been sporadically seen in 3D reconstruction literature, their importance to \revise{reconstruction generalization} has never been identified and evaluated as in our work. \revise{Previous methods have used silhouette loss as supervision mainly in weakly-supervised or self-supervised 3D reconstruction settings, where ground-truth shapes are unavailable. In our paper, we argue that even when ground-truth shapes are available, silhouette loss can still be used as an additional supervision and regularization technique during training, which complements CD loss. Our motivation for using the multi-view silhouette loss is to find a suitable regularization method to improve model generalization. By incorporating this additional loss into the training process, we have achieved more robust and effective results.} For example, sihouette loss is only adopted by unsupervised 3D reconstruction~\cite{pytorch3d,SoftRasterizer,SemanticConsistency,CSMR} as the second choice when 3D shapes are agnostic, while we point out its advantages for supervised learning in this paper.

Our main contributions are summarized as follows.
\textbf{First}, we develop an end-to-end two-stage framework for \revise{single-view 3D mesh reconstruction on seen and unseen categories} by disentangling the task into point cloud generation and surface recovery, which simplifies the problem and alleviates overfitting on seen object categories. \textbf{Second}, we argue that the coarse global feature cannot guide the accurate generation and tend to overfit the training set. To this end, we introduce the local feature sampling strategy in both 2D and 3D spaces and demonstrate the benefits for \revise{reconstruction generalization}. \revise{While previous methods have utilized 2D local features for reconstruction, we want to emphasize that we have demonstrated the superior performance of 2D local features for model generalization. In addition, we have extended this idea into 3D space by introducing 3D local features in our tasks. This approach has enabled us to achieve better results and improve the overall effectiveness of our methods.} \textbf{Third}, we regularize reconstruction by a multi-view silhouette loss to measure surface reconstruction quality within 2D space, which was ignored by point-to-point distances, and furthermore improves the performance with regard to the Chamfer Distance and Earth Mover Distance.  \textbf{Fourth}, we compare the generalization capacity of our model with the existing single-view 3D mesh reconstruction methods on novel objects. The experimental results on ShapeNet.V1~\cite{shapenet} and Pix3D~\cite{pix3d} demonstrate the significant improvement by our \name\ on various metrics.
 %\end{itemize}
%
\begin{figure*}[t]
  \centering
  \includegraphics[width=1.0\linewidth]{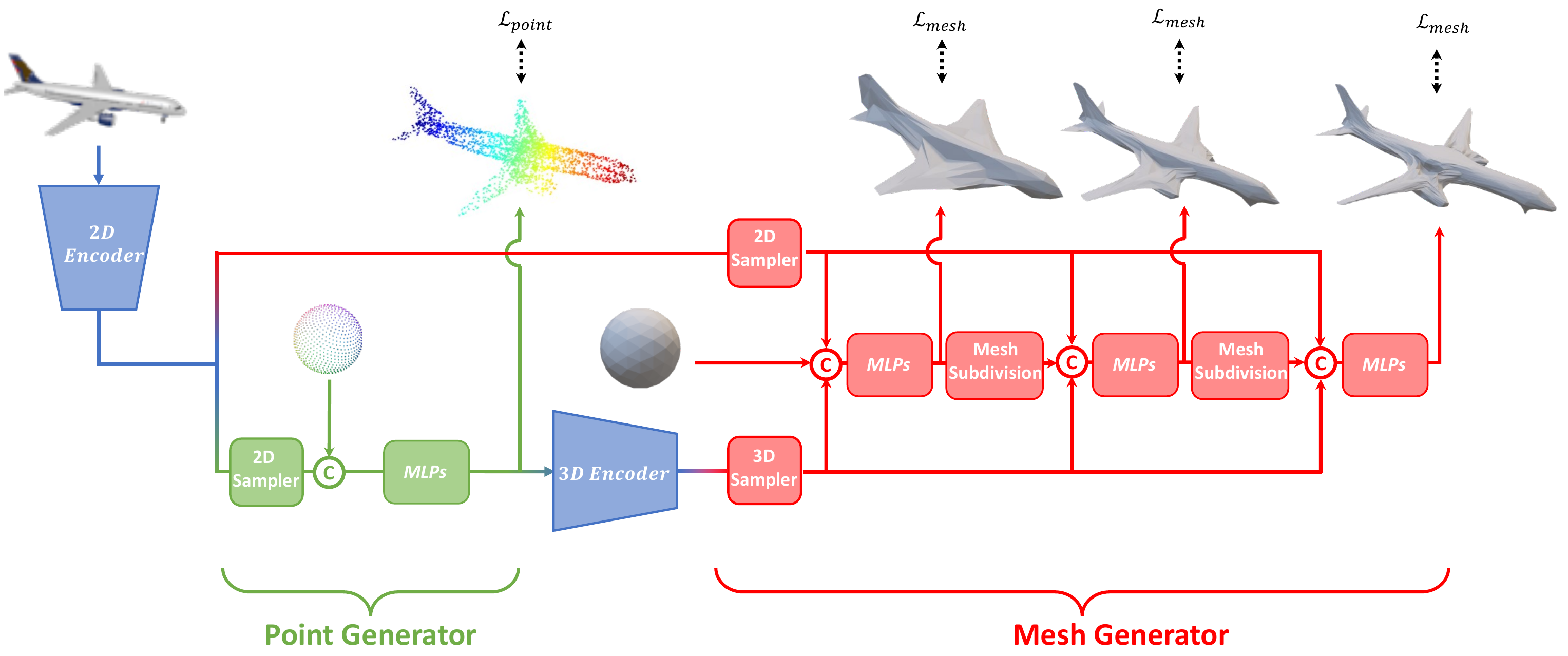}
%   \vspace{-2mm}
  \caption{Network architecture of the proposed \name. The framework includes two main modules: Point Cloud Generator (green) and Mesh Generator (red). Both generators sample local features by point/vertex coordinates by 2D and 3D samplers, respectively. The mesh generator employs a coarse-to-fine process to improve mesh generation iteratively.}
  \label{fig:main}
%   \vspace{-6mm}
\end{figure*}

\section{Related work}
\noindent\textbf{Single-view 3D Reconstruction.}~~Single-view 3D reconstruction is a challenging task. Accurate reconstruction requires the integration of strong geometric priors about our 3D world, which are, however, limited in the wild scenario~\cite{ShapefromShading1981,ShapefromShading_oneview,Local,Pentland1988ShapeIF}. Learning based methods, therefore, become dominant in this field due to their robustness and accessibility. According to the employed 3D representations, the deep learning based methods can be classified into voxel based~\cite{3DR2N2,occnet,richter2018matryoshka,CoReNet,Probabilistic_Latent_Space}, point cloud based~\cite{CD_EMD_LOSS,han2020drwr,chao2021unsupervised}, mesh based~\cite{Neural3DMeshRenderer,SoftRasterizer,InterpolationRenderer,image2mesh,CSMR,MeshRCNN,ShelfSupervised,SemanticConsistency,Total3D}, and implicit function~\cite{ImplicitFields,PIFu,PIFuHD} based frameworks. Among them, mesh reconstruction~\cite{pixel2mesh,atlasnet,tmn} is most related to our work. The majority of the existing single-view 3D mesh reconstruction methods adapts the encoder-decoder framework, where the encoder extracts perceptual features from the input image, and the decoder deforms a template (squire or sphere) to the target 3D shape. Wang \etal~\cite{pixel2mesh} firstly applied deep learning networks to this task, where VGG network~\cite{vgg} was used as the encoder and a graph convolutional network (GCN) was used as the decoder. Groueix \etal~\cite{atlasnet} represented a 3D shape as a collection of parametric surface elements to flexibly represent shapes with arbitrary topology. Pan~\etal~\cite{tmn} focused on the topology changes and proposed a topology modification network by adaptively deleting faces. It is noteworthy that these methods are trained and evaluated on the same object categories. 
\revise{Recent research has also developed methods for 3D reconstruction from image collections without 3D supervision. This has been achieved by utilizing differentiable rendering to supervise the learning of reconstruction. For example, Kanazaw~\etal~\cite{CSMR} propose a method that reconstructs the underlying shape by learning deformations on top of a category-specific mean shape. Lin~\etal~\cite{sdfsrn} develop a differentiable rendering formulation for learning signed distance functions as implicit 3D shape representations, which overcomes topological restrictions. Duggal~\etal~\cite{tars3D} learn both the deformation and implicit 3D shape representations, enabling reconstruction in the category-specific canonical space. Vasudev~\etal~\cite{ss3d} expand category-specific models into cross-category models by implementing distillation. Their training process is divided into three steps: pre-training, self-training, and distilling. Noticing the success of 2D supervision for reconstruction, we introduce the differentiable rendering into supervised learning (ground-truth 3D shapes are available) as an additional supervision on top of the commonly used 3D distance supervision, \ie, Chamfer Distance, and demonstrate its advantages for supervised learning.}

\noindent\textbf{Generalized Single-view 3D Reconstruction.}~~There are also a few research works catering for the generalization capacity of 3D reconstruction models. To the best of our knowledge, there are only three voxel based works~\cite{genre,GSIR,apple_wacv} and one signed distance function based work~\cite{sdfnet} exploring novel class object reconstruction, elaborated as follows. Zhang \etal~\cite{genre} pioneered the generalized single-view voxel reconstruction, where the 2D-3D mapping was decomposed into 2D-2.5D-3D mappings with the use of depth and normals as intermediate representations. Wang \etal~\cite{GSIR} followed that work, and more importantly, introduced shape interpretation into reconstruction and jointly learned interpretation and reconstruction to capture more generic geometry. Bautista \etal~\cite{apple_wacv} studied feature description bias for generalization and emphasized the reconstruction from multiple views. Thai \etal~\cite{sdfnet} shared the similar intermediate representation as~\cite{genre,GSIR} but transferred  the pipeline into the signed distance function by replacing the decoder with the conditional batch normalization network. Besides, there are two papers related to few-shot single-view 3D reconstruction~\cite{few_shot_voxel_2019,few_shot_voxel_2020}, which 
%also explore the model generalization on novel objects, but 
rely on additional 3D inputs, \ie, support shapes.

\section{Methods}
\label{sec:method}
In this section, we present our \name~model that consists of a point cloud generator and a mesh generator in sequence as shown in \cref{fig:main}. Specifically, both of the two generators move the points by predicting per-point offset under the guidance of local features. The local 2D and 3D features are generated by 2D and 3D samplers to enhance model understanding to object local geometry rather than object category, which will be illustrated in the \cref{sec:sample}. To complement 3D point distance in surface supervision, our pipeline differs from previous methods in introducing additional supervision by our proposed multi-view silhouette loss, which will be discussed in \cref{sec:loss}.

%===================================
\subsection{Two-stage reconstruction}
\label{sec:2stage}
\name\ disentangles the complicated image-to-mesh generation into two simpler generation tasks, \ie, point cloud generation and surface recovery, through a Point Cloud Generator and a Mesh Generator, elaborated as follows.

\noindent\textbf{Point Cloud Generator.}~~The point cloud generator takes the input 2D image as the guidance and transforms the point cloud template sampled from a unit sphere into the target point cloud $P\in \mathbb{R}^{p\times 3}$, where $p$ is the number of points. Specifically, the input 2D image is passed to a 2D encoder, such as ResNet, to extract feature maps $F\in \mathbb{R}^{c\times h\times w}$, upon which, a 2D sampler is applied to sample local features. These local features are then sent into a multi-layer perceptron (MLP) to predict per-point offset. The whole process could be formulated as:
%To obtain the intermediate point cloud representation, we transform the template points (points randomly sampled from unit sphere) into target points $P\in \mathbb{R}^{p\times 3}$. The point generator consists of a 2D sampler for local feature sampling from image feature maps $F\in \mathbb{R}^{c\times h\times w}$ and a multi-layer perceptron (MLP) to predict per-point offset, which is modeled as
\begin{equation}
    \hat{\boldsymbol{p_{i}}} = \boldsymbol{p_i} + MLP([\boldsymbol{p_i}|f_i]),
\end{equation}
where $\boldsymbol{p_i}$ and $\hat{\boldsymbol{p_{i}}}$ indicate the $i$-th point before and after the transformation, $f_i$ is the local feature of $p_i$ sampled by the 2D sampler, and $[\cdot|\cdot]$ denotes a concatenation operation. The point cloud generator is supervised by Chamfer Distance between the generated and the ground-truth point clouds.

 \noindent\textbf{Mesh Generator.}~~Taking the 2D image features and the 3D point cloud features as guidance, our mesh generator reconstructs the target object $M=(V,E)$, where $V\in \mathbb{R}^{v\times 3}$ and $E\in \mathbb{R}^{e\times 2}$ denote the vertices and the edges, respectively. The reconstruction is achieved by moving the vertices of a unit sphere mesh template towards the ground-truth vertices while maintaining the edge connections. Specifically, the intermediate point cloud output by the point cloud generator is sent into a 3D encoder, such as PointNet$^{++}$~\cite{pointnet++}, to extract feature groups $G\in \mathbb{R}^{n\times c}$ as the 3D information source of the mesh generator. We then sample the per-vertex 2D feature $f_i$ from the 2D image feature maps $F\in \mathbb{R}^{c\times h\times w}$, and the per-vertex 3D feature $g_i$ from the 3D feature groups $G\in \mathbb{R}^{n\times c}$ ($n$ is the number of groups),
 %the 3D feature groups $G\in \mathbb{R}^{n\times c}$, we sample the per-vertex 2D feature $f_i$ and 3D feature $g_i$, 
 and then concatenate them with the template coordinate $\boldsymbol{v_i}$ to recover the final mesh. Please see~\cref{sec:sample} for feature sampling. Our mesh generation is a coarse-to-fine process composed of a sequence of modules for refinement and subdivision to decrease the overfitting and intersection, similar to Pixel2Mesh~\cite{pixel2mesh}. Specifically, in the $i$-th module, the current predicted mesh $M_i$ is firstly refined according to the output of the $(i-1)$-th module through an MLP, and then subdivided by breaking each triangle face into four faces via adding three vertices at the mid-point of the triangle edges. The updated mesh $M_i$ is then output by the $i$-th module and used as the input of the $(i+1)$-th module for another round of refinement. 
 %Our mesh generation is a coarse-to-fine process by refinement and subdivision in sequence to decrease the over-fitting and intersection, similar to Pixel2Mesh~\cite{pixel2mesh}. Specifically, in the $i$-th deformation step, a MLP move the vertices of the input template $M_i$, named refinement, and then the mesh subdivision divides each triangle face into four faces by adding three vertices at the mid-point of the triangle edges. After mesh subdivision, the output of $i$-th module $M_i$ was taken as the input template of the next step to get $M_{i+1}$.  
 Our mesh generator predicts the per-vertex offset through an MLP so that after deformation the $i$-th vertex $v_i$ is,
\begin{equation}
    \hat{\boldsymbol{v_{i}}} = \boldsymbol{v_i} + MLP([\boldsymbol{v_i}|f_i|g_i]),
\end{equation}
where $\boldsymbol{v_i}$ and $\hat{\boldsymbol{v_{i}}}$ denote the $i$-th vertex before and after the deformation, $f_i$ and $g_i$ denote the local features by 2D sampling and 3D sampling, respectively, and $[\cdot|\cdot|\cdot]$ is a concatenation operation. 
The mesh losses in ~\cref{sec:overall_loss} are applied on all module outputs $\{M_1, M_2, ..., M_n\}$ for supervision.

% \begin{figure}[t]
%   \centering
%   \begin{subfigure}{0.4\textwidth}
%     \includegraphics[width=0.2\textwidth]{2dsampler.pdf}
%     \caption{2D Sampler}
%     \label{fig:2dsampler}
%   \end{subfigure}
%   \hfill
%   \begin{subfigure}{0.4\textwidth}
%     \includegraphics[width=0.2\textwidth]{3dsampler.pdf}
%     \caption{3D Sampler}
%     \label{fig:3dsampler}
%   \end{subfigure}
% \caption{Local feature Sampler. (a) 2D local features are sampled by 2D projection according to camera intrinsic matrix, camera pose and point/vertex coordinates. (b) 3D local features are sampled by nearest neighbor searching according to vertex coordinates and feature group centroids.}
% \label{fig:sampler}
% \vspace{-5mm}
% \end{figure}

\begin{figure*}[t]
\centering
\subfloat[]{\includegraphics[width=2.5in]{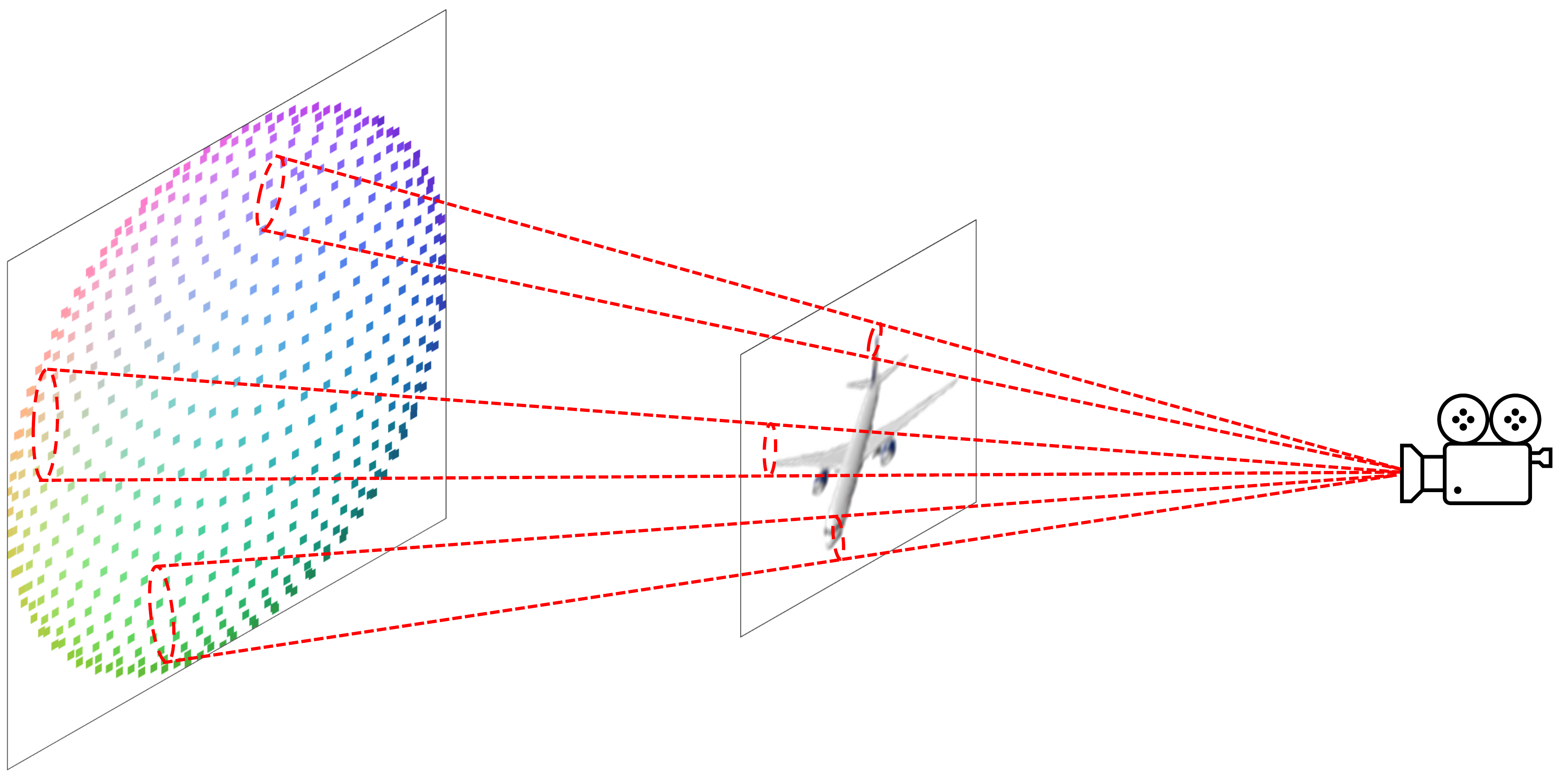}%
\label{2dsampler}}
\hfil
\subfloat[]{\includegraphics[width=2.5in]{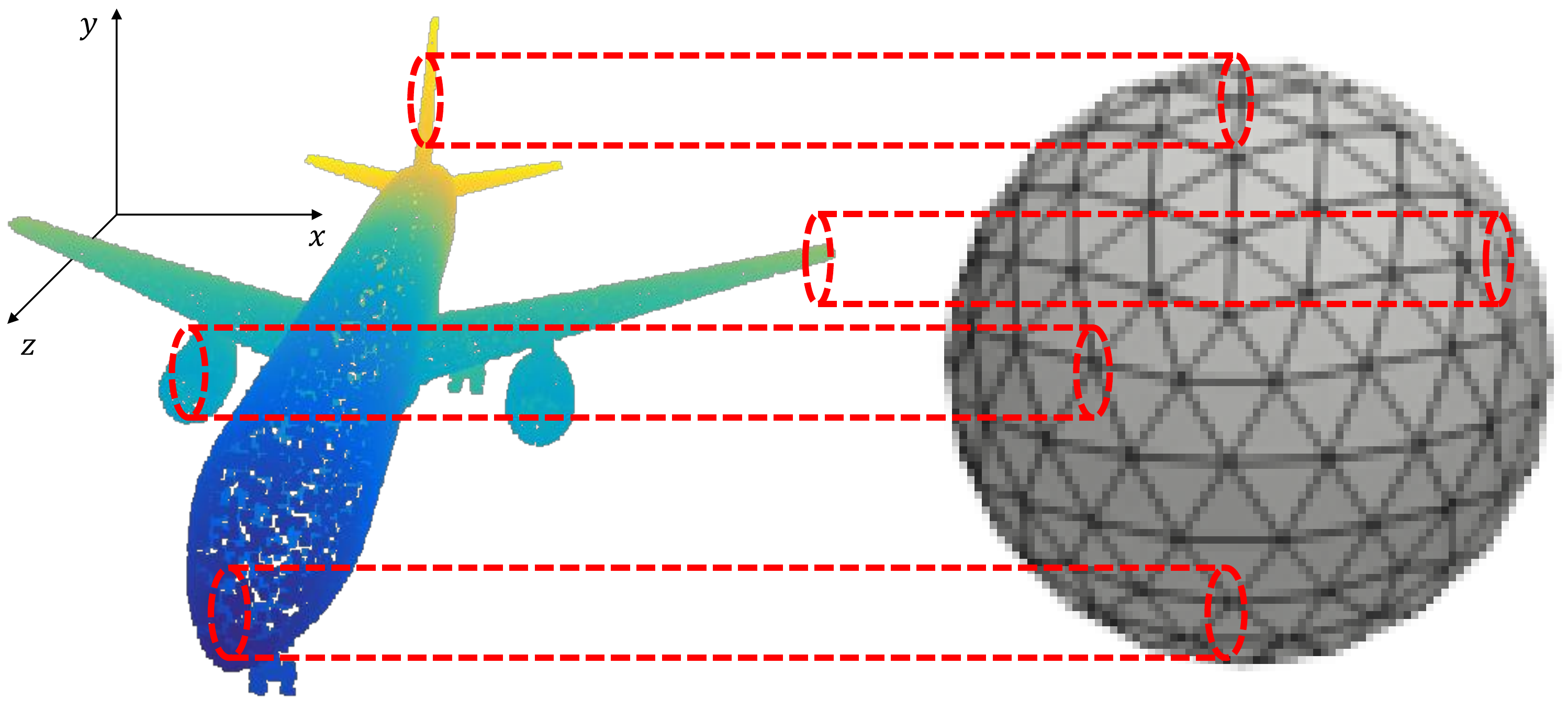}%
\label{3dsampler}}
\caption{Local feature Sampler. \revise{(a) 2D local features are sampled by 2D projection according to camera intrinsic matrix, camera pose and point/vertex coordinates.} (b) 3D local features are sampled by nearest neighbor searching according to vertex coordinates and feature group centroids.}
\label{fig:sampler}
\end{figure*}

\subsection{Feature Sampling}
\label{sec:sample}
Local features associated with each 3D point are extracted by sampling the RGB image features $F\in \mathbb{R}^{c\times h \times w}$ through the 2D sampler or sampling the intermediate point cloud features $G\in \mathbb{R}^{n\times c}$ through the 3D sampler.
  
  %The RGB image features $F\in \mathbb{R}^{c\times h \times w}$ or the intermediate point cloud features $G\in \mathbb{R}^{n\times c}$ are initially encoded by 2D or 3D encoders, such as ResNet or PointNet++~\cite{pointnet++}. Following that, the 2D or 3D sampler is deployed to obtain local features $f_i\in \mathbb{R}^{c}$ or $g_i\in \mathbb{R}^{c}$ according to vertex $\boldsymbol{x_i}$ as illustrated in \cref{fig:sampler}.
\revise{
 \noindent\textbf{2D sampler.}~~For each point $\boldsymbol{x_i}=(x_i, y_i, z_i)$ from the point cloud $P\in\mathbb{R}^{p\times 3}$ or the vertices $V\in\mathbb{R}^{v\times 3}$, we calculate its 2D projection on the image plane and sample local features from the image feature maps by bilinear interpolation. Specifically, for 2D projection, we first transfer the point coordinates from the world coordinate $\boldsymbol{x_i}$ into the camera coordinate $\boldsymbol{x_i}^{'}$ using camera intrinsic matrix $K\in \mathbb{R}^{3\times 4}$, and then calculate the point position $\boldsymbol{p_i}$ on the image plane. That is, $(x'_i, y'_i, z'_i, 1)=[\boldsymbol{x_i^{'}}|1] = K\left[R|T\right]\left[\boldsymbol{x_i}|1\right] $
  and $\boldsymbol{p^{2d}_i}=(\frac{x_i^{'}}{z_i^{'}},\frac{y_i^{'}}{z_i^{'}})$,
% \begin{equation}
%     [\boldsymbol{x_i^{'}}|1] = K\left[R|T\right]\left[\boldsymbol{x_i}|1\right],\ \ \ \boldsymbol{p_i}=(\frac{x_i^{'}}{z_i^{'}},\frac{y_i^{'}}{z_i^{'}}),
% \end{equation}
% \begin{equation}
% \begin{pmatrix}
% \frac{x'_i}{z'_i}\\
% \frac{y'_i}{z'_i}\\
% 1
% \end{pmatrix} \sim \begin{pmatrix}
% x'_i\\
% y'_i\\
% z'_i\\
% 1
% \end{pmatrix}=\begin{pmatrix}
% f & 0 & 0 & 0 \\
% 0 & f & 0 & 0 \\
% 0 & 0 & 1 & 0 \\
% 0 & 0 & 0 & 1
% \end{pmatrix}\begin{pmatrix}
% r_{11} & r_{12} & r_{13} & t_1 \\
% r_{11} & r_{12} & r_{13} & t_1 \\
% r_{11} & r_{12} & r_{13} & t_1 \\
% 0 & 0 & 0 & 1
% \end{pmatrix}\begin{pmatrix}
% x_i\\
% y_i\\
% z_i\\
% 1
% \end{pmatrix}
% \end{equation}
where $R\in \mathit{SO}(3)$ and $T\in \mathbb{R}^{3\times 1}$ are the rotation and the translation matrices. After projection, the local features are interpolated from the four nearby pixels around the position $\boldsymbol{p^{2d}_i}$ on the 2D feature maps $F\in \mathbb{R}^{c\times h\times w}$. The 2D local features corresponding to the point $\boldsymbol{x_i}$ are defined as 
\begin{equation}
  f_i = \sum_{j\in\mathcal{N}(\boldsymbol{p^{2d}_i})}\omega_{j}f_j,
\end{equation}
where $\mathcal{N}(p^{2d}_i)$ denotes the nearest four corner pixels of the position $\boldsymbol{p^{2d}_i}$; $f_j\in \mathbb{R}^{c}$ is the feature at the position $\boldsymbol{p^{2d}_j}$ extracted from $F\in \mathbb{R}^{c\times h \times w}$, and the bilinear interpolation weights $\omega_{j}$ are calculated according to the pixel position $\boldsymbol{p^{2d}_j}$ and the projected point position $\boldsymbol{p^{2d}_i}$.
}

  \noindent\textbf{3D sampler.}~~A given 3D point $\boldsymbol{x_i}$ is assigned 3D local features by nearest neighbor searching on the 3D feature groups $G\in\mathbb{R}^{n\times c}$ extracted by the 3D encoder from the intermediate point cloud. Each feature group encodes the local points within the sphere with radius $r$. During encoding, to guarantee translation invariant and capture relative correlation, the coordinates of points in a local region are firstly translated into a local frame relative to the centroid point: $\boldsymbol{x_i}=\boldsymbol{x_i}-\boldsymbol{\bar{x}}$ for $i=1,2, ..., p$, where $\boldsymbol{\bar{x}}$ is the coordinate of the centroid and $p$ is the point number belonging to the group. For more details about the feature groups, please refer to~\cite{pointnet++}. Given the 3D feature groups with the group centroids $\{(g_1, \boldsymbol{\bar{x}_1}), (g_2, \boldsymbol{\bar{x}_2}), ..., (g_n, \boldsymbol{\bar{x}_n})\}$ and the given point $\boldsymbol{x_i}$, the 3D local feature sampling is defined as,

  \begin{equation}
    g_i = \min_{\boldsymbol{\|\boldsymbol{x_i}-\bar{\boldsymbol{x}}_j}\|_2^2}\{g_j\}, \ \ j=1,2, ...,n
\end{equation}
  where $\boldsymbol{x_i}$ is query point and $\bar{\boldsymbol{x}}_j$ is the $j$-th group centroid. Note that the 3D sampler is only deployed at the second stage, \ie, mesh generation.

\subsection{Multi-view Silhouette Loss}
\label{sec:loss}
To introduce face quality supervision and regularization into training, we render the predicted mesh and the ground-truth mesh into silhouettes by multiple virtual cameras and calculate the Intersection over Union (IoU) between the corresponding predicted and ground-truth binary masks. The multi-view silhouette loss is formulated as,\\
\begin{equation}
    \mathcal{L}(P, Q)=\frac{1}{|V|}\sum _{V}^{|v|}\sum _{i}^{hw}1-\frac{{p_{i}^{v}q_i^v}}{p_i^v+ q_i^v - p_{i}^{v}q_i^v},
\end{equation}
where $P,Q\in\mathbb{R}^{v\times h\times w}$ are the predicted and the ground-truth silhouettes, respectively, from the $v$ camera views by differentiable rendering~\cite{pytorch3d}. 

 \begin{figure}[t]
    \begin{center}
    \includegraphics[width=0.35\textwidth]{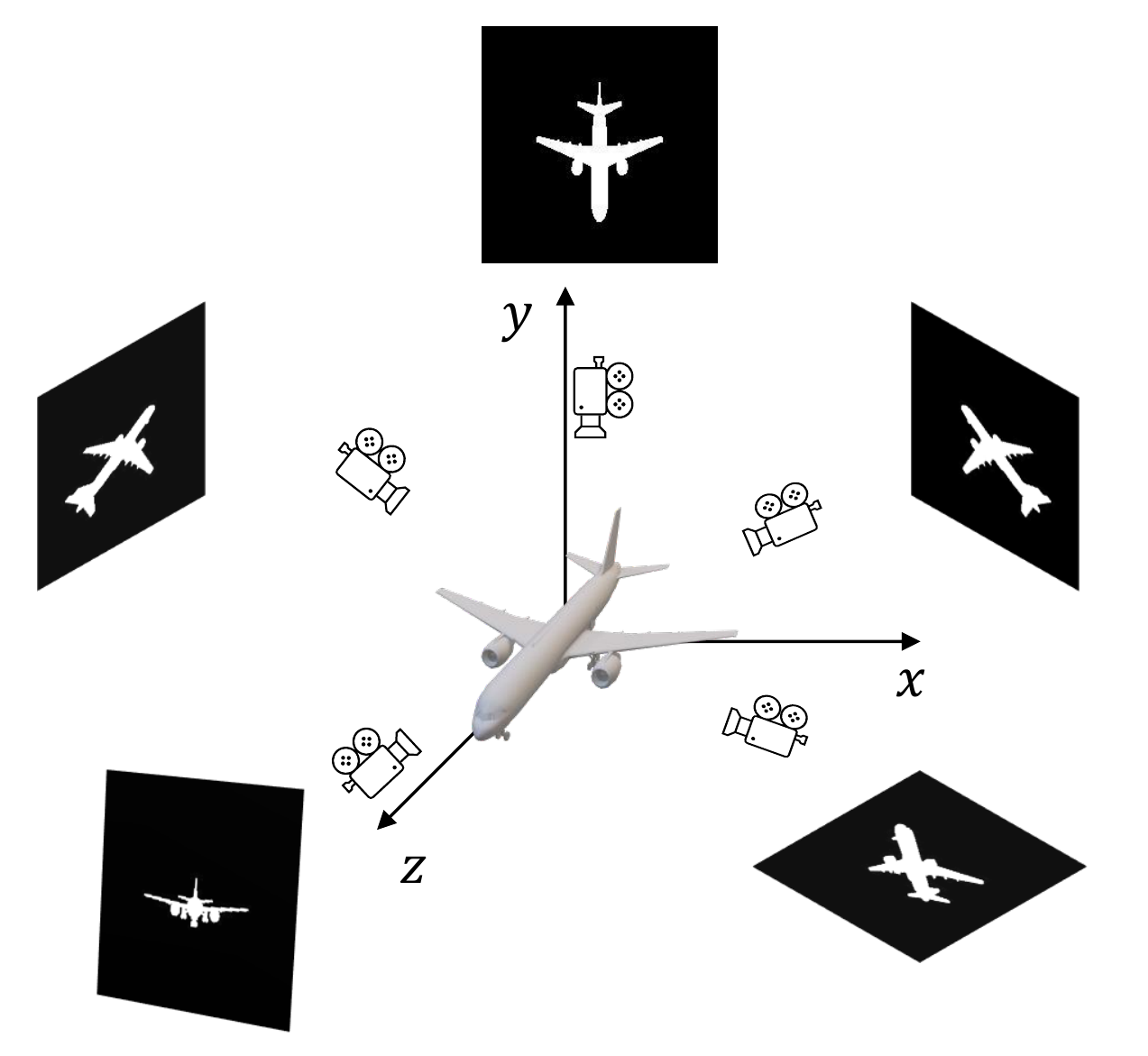}
  \end{center}
%   \vspace{-20pt}
  \caption{Multi-view silhouette rendering. Multiple cameras are set at different positions to render silhouette maps.}
% \vspace{-20pt}
\end{figure}

\subsection{Overall Loss}
\label{sec:overall_loss}
In addition to our proposed multi-view silhouette loss, we adopt the widely used losses in 3D space, \ie, Chamfer Distance $\mathcal{L}_{chamfer}$ and the normal loss $\mathcal{L}_{normal}$. The Chamfer Distance is to measure the distance between two point sets, and the normal loss minimizes the normal directions between predicted and ground-truth mesh. The two losses are defined as follows:
\begin{equation*}
    \mathcal{L}_{chamfer}=\sum_{p}\min_q\| p-q\| _2^2+\sum_{q}\min_p\| q-p\| _2^2
\end{equation*}
\begin{equation}
    \mathcal{L}_{normal}=\sum_{p}\min_q\|n_p-n_q\| _2^2+\sum_{q}\min_p\|n_q-n_p\| _2^2,
\end{equation}
where $p$ and $q$ are the points from the predicted and the ground-truth point sets, $n_p$ and $n_q$ are the normals corresponding to the points $p$ and $q$. The points and normals are randomly sampled from the surface of the generated and ground-truth mesh $M=(V,E)$. Moreover, to improve mesh quality, we also apply the edge length loss $\mathcal{L}_{edge}$ and the vertex move loss $\mathcal{L}_{move}$ to penalize too long edges and dramatic moving for vertices during deformation, as follows:
\begin{equation}
    \mathcal{L}_{edge}=\sum_{e\in E}\| \boldsymbol{e_1}-\boldsymbol{e_2}\| _2^2, \ \ \  \mathcal{L}_{move}=\sum_{\boldsymbol{v}\in V}\| \boldsymbol{\hat{v}}-\boldsymbol{v}\| _2^2,
\end{equation}
where $\boldsymbol{e_1}$ and $\boldsymbol{e_2}$ are the two vertices of an edge, and $\boldsymbol{v}$ and $\boldsymbol{\hat{v}}$ indicate the vertices before and after the deformation. 

Putting it together, the overall loss function is,
\begin{equation}
    \mathcal{L}=\mathcal{L}_{point}+\mathcal{L}_{mesh},
\end{equation}
where $\mathcal{L}_{point}$ is Chamfer Distance $\mathcal{L}_{chamfer}$, and the $\mathcal{L}_{mesh}$ is defined as,
\begin{align}
    \mathcal{L}_{mesh}=&\mathcal{L}_{chamfer}+\lambda_{normal}\mathcal{L}_{normal}+\lambda_{IoU}\mathcal{L}_{IoU} \nonumber\\ &+\lambda_{edge}\mathcal{L}_{edge}+\lambda_{move}\mathcal{L}_{move},
\end{align}
% \begin{align*}
%     \mathcal{L}=&\mathcal{L}_{point}+\mathcal{L}_{mesh} \\
%     \mathcal{L}_{mesh}=&\mathcal{L}_{chamfer}+\lambda_{normal}\mathcal{L}_{normal}+\lambda_{IoU}\mathcal{L}_{IoU} \\ &+\lambda_{edge}\mathcal{L}_{edge}+\lambda_{move}\mathcal{L}_{move}, \\
%     \mathcal{L}_{point}=&\mathcal{L}_{chamfer},
% \end{align*}
where the hyper-parameters $\lambda_{normal}$, $\lambda_{IoU}$, $\lambda_{edge}$, and $\lambda_{move}$ are simply set to balance these loss terms.

\subsection{Implementations}
We adopt ResNet18~\cite{resnet} and PointNet$^{++}$~\cite{pointnet++} as our 2D and 3D encoders. To obtain the local features with different sizes of the receptive field, the 2D feature sampling is conducted on the feature maps from ResBlock 2, 3, and 4, and the 3D feature sampling is on the feature groups from the grouping layer 1, 2, and 3. Final pooling layers and fully connected layers in both are not used. The resolution of the input images is 224 × 224. We set the batch size as $12$, the total training epoch as $250$, and the learning rate as $1\times 10^{-3}$ with the decay of $0.3$ at the $100$-th, $200$-th, and $230$-th epoch. 
The values of hyperparameters used in the overall loss are set as $\lambda_{normal}=10^{-3}$, $\lambda_{edge}=10^{-1}$, $\lambda_{move}=10^{-3}$, $\lambda_{IoU}=10^{-2}$ without any elaborated adjustments. The multi-view silhouette loss $\mathcal{L}_{IoU}$ is only applied on the final mesh after 100 epochs due to time cost. 
\revise{At the first stage, we randomly sample $p=2500$ points from the unit sphere as the input $P\in \mathbb{R}^{p\times 3}$ to the point cloud generator. At the second stage, the input to the mesh generator is the icosphere $M=(V, E)$ with $V=162$ vertices. The outputs of the mesh generator are $\{M_1, M_2, M_3\}$ with $V_1=162$ vertices, $V_2=642$ vertices, and $V_3=2562$ vertices, in sequence. The feature maps extracted by ResNet18 are $F_1\in \mathbb{R}^{128\times 28\times 28}$, $F_2\in \mathbb{R}^{256\times 14\times 14}$, and $F_3\in \mathbb{R}^{512\times 7\times 7}$. The feature groups extracted by PointNet++ are $G_1\in \mathbb{R}^{128\times 512}$, $G_2\in \mathbb{R}^{256\times 128}$, and $G_3\in \mathbb{R}^{512\times 1}$. The sampled 2D feature $f_i\in \mathbb{R}^{c}$ and 3D feature $g_i\in \mathbb{R}^{c}$ are with the dimension $c=128+256+512=896$ both. Regarding the multi-view silhouette loss, we sample 24 viewpoints around the objects with elevation $\{-45, 0, 45\}$, azimuth $\{0, 45, 90, 135, 180, 225, 270, 315\}$, and random distance uniform distribution $(1.1375, 1.6625)$. We only sample 8 masks to calculate multi-view silhouette loss during training due to memory constraint, and utilize the 24 views to for evaluation. We apply the MLPs to all vertices and the weights of the MLPs are shared.}

\begin{table*}[!h]
\caption{Performance comparison on ShapeNet under Chamfer Distance (CD), Earth Mover Distance (EMD), Multi-view Silhouette IoU (IoU), Normal Consistency (NC), and F-score (F). Best results are bolded. The symbols \textit{v} and  \textit{o} represent the object-centered and the viewer-centered coordinates, respectively.}
\label{table:comparison}
\centering
\resizebox{0.95\textwidth}{!}{
\begin{tabular}{c|ccccc|ccccc}
\hline
\multirow{2}{*}{Methods} & \multicolumn{5}{c|}{Base Classes} & \multicolumn{5}{c}{Novel Classes} \\ \cline{2-11} 
 & \multicolumn{1}{c|}{\makecell[c]{CD$\downarrow$\\$\times10^{-3}$}} & \multicolumn{1}{c|}{\makecell[c]{EMD$\downarrow$\\$\times10^{-2}$}} & \multicolumn{1}{c|}{\makecell[c]{IoU$\uparrow$\\$\times10^{-2}$}} & \multicolumn{1}{c|}{\makecell[c]{NC$\downarrow$\\$\times10^{-1}$}} & \multicolumn{1}{c|}{\makecell[c]{F-score$\uparrow$\\$\times10^{-2}$}} & \multicolumn{1}{c|}{\makecell[c]{CD$\downarrow$\\$\times10^{-3}$}} & \multicolumn{1}{c|}{\makecell[c]{EMD$\downarrow$\\$\times10^{-2}$}} & \multicolumn{1}{c|}{\makecell[c]{IoU$\uparrow$\\$\times10^{-2}$}} & \multicolumn{1}{c|}{\makecell[c]{NC$\downarrow$\\$\times10^{-1}$}} &
 \multicolumn{1}{c}{\makecell[c]{F-score$\uparrow$\\$\times10^{-2}$}}\\ \hline

AtlasNet~\cite{atlasnet} & \multicolumn{1}{c|}{5.35} & \multicolumn{1}{c|}{7.56} & \multicolumn{1}{c|}{78.63} & \multicolumn{1}{c|}{4.72} & \multicolumn{1}{c|}{52.84} & \multicolumn{1}{c|}{23.91} & \multicolumn{1}{c|}{11.60} & \multicolumn{1}{c|}{67.17} & \multicolumn{1}{c|}{7.14} & \multicolumn{1}{c}{39.65}\\
Pixel2Mesh~\cite{pixel2mesh} & \multicolumn{1}{c|}{7.55} & \multicolumn{1}{c|}{8.87} & \multicolumn{1}{c|}{77.14}  & \multicolumn{1}{c|}{5.30} & \multicolumn{1}{c|}{45.71} & \multicolumn{1}{c|}{11.81} & \multicolumn{1}{c|}{11.51} & \multicolumn{1}{c|}{72.73} & \multicolumn{1}{c|}{6.52} & \multicolumn{1}{c}{39.06} \\ 
%  Pixel2Mesh$^+$ (o) & \multicolumn{1}{c|}{4.47} & \multicolumn{1}{c|}{5.69} & \multicolumn{1}{c|}{82.69} & \multicolumn{1}{c|}{4.70} & \multicolumn{1}{c||}{56.24} & \multicolumn{1}{c|}{8.01} & \multicolumn{1}{c|}{6.30} & \multicolumn{1}{c|}{78.04} & \multicolumn{1}{c|}{6.14} & \multicolumn{1}{c}{45.87}\\ 
 OccNet~\cite{occnet} & \multicolumn{1}{c|}{8.63} & \multicolumn{1}{c|}{5.09} & \multicolumn{1}{c|}{73.68}  & \multicolumn{1}{c|}{\textbf{4.41}} & \multicolumn{1}{c|}{41.00} & \multicolumn{1}{c|}{40.48} & \multicolumn{1}{c|}{10.59} & \multicolumn{1}{c|}{65.31} & \multicolumn{1}{c|}{6.60} & \multicolumn{1}{c}{29.32} \\
Mesh R-CNN~\cite{MeshRCNN} & \multicolumn{1}{c|}{6.04} & \multicolumn{1}{c|}{\bf{4.68}} & \multicolumn{1}{c|}{77.25} & \multicolumn{1}{c|}{6.50} & \multicolumn{1}{c|}{47.46} & \multicolumn{1}{c|}{8.84} & \multicolumn{1}{c|}{\bf{5.60}} & \multicolumn{1}{c|}{74.10} & \multicolumn{1}{c|}{7.50} & \multicolumn{1}{c}{40.50}\\ 
TMNet~\cite{tmn} & \multicolumn{1}{c|}{6.07} & \multicolumn{1}{c|}{5.85} & \multicolumn{1}{c|}{78.81} & \multicolumn{1}{c|}{4.68}& \multicolumn{1}{c|}{53.12} & \multicolumn{1}{c|}{32.79} & \multicolumn{1}{c|}{11.40} & \multicolumn{1}{c|}{66.23} & \multicolumn{1}{c|}{7.20} & \multicolumn{1}{c}{37.51}\\ 
SDFNet~\cite{sdfnet} & \multicolumn{1}{c|}{13.22} & \multicolumn{1}{c|}{7.41} & \multicolumn{1}{c|}{73.73} & \multicolumn{1}{c|}{4.79} & \multicolumn{1}{c|}{34.37}& \multicolumn{1}{c|}{23.20} & \multicolumn{1}{c|}{9.34} & \multicolumn{1}{c|}{68.51} & \multicolumn{1}{c|}{6.20} & \multicolumn{1}{c}{25.09} \\ \hline

% Ours (v)  & \multicolumn{1}{c|}{6.04} & \multicolumn{1}{c|}{6.54} & \multicolumn{1}{c|}{74.70} & \multicolumn{1}{c|}{4.97} & \multicolumn{1}{c||}{45.66} & \multicolumn{1}{c|}{8.54} & \multicolumn{1}{c|}{6.99} & \multicolumn{1}{c|}{74.02} & \multicolumn{1}{c|}{6.24} & \multicolumn{1}{c}{39.15} \\ 
Ours (v)  & \multicolumn{1}{c|}{4.18} & \multicolumn{1}{c|}{5.54} & \multicolumn{1}{c|}{83.91} & \multicolumn{1}{c|}{4.67} & \multicolumn{1}{c|}{57.23} & \multicolumn{1}{c|}{6.71} & \multicolumn{1}{c|}{6.16} & \multicolumn{1}{c|}{79.85} & \multicolumn{1}{c|}{5.96} & \multicolumn{1}{c}{47.61} \\
Ours (o) & \multicolumn{1}{c|}{\bf{3.96}} & \multicolumn{1}{c|}{5.36} & \multicolumn{1}{c|}{\bf{85.85}} & \multicolumn{1}{c|}{4.51} & \multicolumn{1}{c|}{\bf{59.57}} & \multicolumn{1}{c|}{\bf{6.69}} & \multicolumn{1}{c|}{5.96} & \multicolumn{1}{c|}{\bf{81.50}} & \multicolumn{1}{c|}{\bf{5.80}} & \multicolumn{1}{c}{\bf{50.09}}\\ \hline
\end{tabular}
}
\end{table*}

\begin{table*}[!h]
\caption{Quantitative comparison on Pix3D, with the model trained on ShapeNet, under Chamfer Distance (CD), Earth Mover Distance (EMD), Multi-view Silhouette IoU (IoU), Normal Consistency (NC), and F-score. 
% \textit{NC} refers to Normal Consistency multiplied by $10$. The threshold of F-score is $10^{-3}$. 
The symbols \textit{v} and \textit{o} represent the viewer-centered and the object-centered coordinates.}
\centering
\resizebox{0.7\textwidth}{!}{
    \begin{tabular}{@{}c|c|c|c|c|c@{}}
    \hline
     Methods & CD$\downarrow\times10^{-3}$ & EMD$\downarrow\times10$ & IoU$\uparrow\times10^{-2}$ & NC$\downarrow\times10$ & F-score$\uparrow\times10^{-2}$ \\
    \hline
    AtlasNet~\cite{atlasnet} & 59.75  & 21.41 & 52.14  & 8.72 & 15.79 \\
     Pixel2Mesh~\cite{pixel2mesh} & 87.44   & 20.79 & 46.49  & 9.16 & 8.96\\
     OccNet~\cite{occnet} & 57.37 & 14.05 & \textbf{55.32} & 7.51 & 14.32\\
     Mesh R-CNN~\cite{MeshRCNN} & 58.74  & 16.84 & 46.95  & 9.27 & 12.97\\
     TMNet~\cite{tmn} & 74.64  & 20.38 & 46.08   & 8.49 & 12.34 \\
     SDFNet~\cite{sdfnet} & 99.82  & 21.28 & 49.58  & 8.98  & 6.49\\ \hline
     Ours (v) & \textbf{30.00}  & 16.22  & 52.38 & \textbf{7.43}  & \textbf{21.24}\\
     Ours (o) & 34.87  & \textbf{13.64}  & 52.03  & 7.62 & 19.33\\
     \hline
    \end{tabular}
    }
    \label{tab:pix3d}
\end{table*}

%===================================================%
\section{Results}
\label{sec:results}
\textbf{Dataset.}~We demonstrate the effectiveness of our model on the ShapeNetCore v1.0~\cite{shapenet} dataset and the Pix3D~\cite{pix3d} dataset. The \textbf{ShapeNet} dataset contains 55 shape classes and we only take 16 classes with relatively large number of objects, and split them into the base classes  and the novel classes. The base classes include \textit{car, chair, monitor, plane, rifle, speaker, table} and \textit{telephone,} and the novel classes include \textit{bench, bus, cabinet, lamp, pistol, sofa, train} and \textit{watercraft.} To reduce the class imbalance, we only randomly sample 200 shapes from each class for testing. Among the 16 classes, the RGB images of 13 classes are provided by ~\cite{3DR2N2}, and the images of the remaining 3 classes are rendered by ourselves using Blender with the consistent rendering setting. The \textbf{Pix3D} dataset only contains 9 classes, 10069 real-world images and 395 unique 3D models. Among the 9 classes, 2 classes, \ie, \textit{tools} and \textit{misc}, only consists of 47 and 68 images. Splitting the remaining 7 classes into base classes and novel classes and training on them is not suitable for our setting due to the diversity and size. Thus, we directly test the generalization of our \name\ and other methods on the Pix3D~\cite{pix3d} dataset by using the model trained on the ShapeNet~\cite{shapenet} without any further training or refinement. Due to the large domain gaps, we randomly sample 100 images from the 7 classes \textit{bed, bookcase, chair, desk, sofa, table, wardrobe} and take them all as novel classes for testing.

\textbf{Evaluation criteria.}~The reconstruction performance is evaluated by five criteria, \ie, Chamfer Distance (CD), Earth Mover Distance (EMD), Multi-view Silhouette Intersection over Union (IoU), Normal Consistency (NC) and F-score. Specifically, the CD and EMD are distance metrics between point sets~\cite{CD_EMD_LOSS}. The F-score is defined as the harmonic mean between the precision and the recall, based on if a prediction/ground-truth point can find any other ground-truth/prediction point within the threshold $\tau=10^{-3}$. The NC measures the cosine error of the prediction normal from its ground-truth~\cite{pixel2mesh,MeshRCNN,whatdolearn}. To calculate the point-based metrics, we first randomly sample 2500 points and normals from each generated surface and its ground-truth, respectively, and then measure the distance upon the two sampled point sets. Since neither CD nor EMD takes into account the surface/mesh connectivity, our proposed IoU loss is used as another evaluation criterion to further account for mesh quality. Specifically, we render the output meshes into silhouettes (binary masks) from different camera views and calculate mean Intersection over Union (IoU) broadly-utilized in segmentation.

\noindent\textbf{Viewer-centered vs Object-centered.} There are two choices of coordinate system: \textbf{viewer-centered}~\cite{pixel2mesh,MeshRCNN,sdfnet} and \textbf{object-centered}~\cite{atlasnet,tmn} coordinates according to pose of reconstruction. The viewer-centered reconstruction improves the generalization on unseen objects but leads to scale-depth ambiguity. The object-centered reconstruction is more friendly to downstream tasks, \eg, analysis, editing, rendering, and generalize to new domains (different camera setting) but relies on camera pose to obtain 2D local features. Accordingly, we provide two variants of our frameworks under both coordinates, respectively. Our viewer-centered variant takes $azimuth = 0$ and $elevation =0$ as camera pose and reconstructs objects aligned with input images. We evaluate all methods with canonical pose by rotation, translation and normalization using ground-truth pose to eliminate depth-scale error.
%For fair comparison, we also provide a modified Pixel2Mesh$^+$ which directly outputs shapes under object-centered coordinates as our local-feature-based object-centered baseline in~\cref{tab:ablaton}, where we also replace the graph convolution with MLPs and deploy better training recipe consistent with our \name.}

\subsection{Quantitative Results.}
We evaluate our model and compare it to five state-of-the-art methods for single-view 3D reconstruction, \ie, Pixel2Mesh~\cite{pixel2mesh}, Mesh R-CNN~\cite{MeshRCNN}, AtlasNet~\cite{atlasnet}, Topology Modification Network (TMNet)~\cite{tmn} and SDFNet~\cite{sdfnet}, while the first four methods are mesh-based reconstruction and SDFNet is a signed distance function method addressing novel classes. Under both the normal and \revise{unseen-category} settings, we use the samples from the base classes for training, whilst the tests are conducted on the seen base classes under the normal setting and on the unseen novel classes. We re-run the shared codes for the four mesh-based methods for a fair comparison, and employ the released SDFNet-Img code~\cite{sdfnet} for SDFNet test.
%while the SDFNet is the SDFNet-Img version reported in~\cite{sdfnet} because they only released code of it.
%All of these methods are trained on the base classes and tested both on the base classes and the novel classes. We report all metrics results for base classes and novel classes. 
Note that Pixel2Mesh~\cite{pixel2mesh}, Mesh R-CNN~\cite{MeshRCNN} and SDFNet~\cite{sdfnet} use the viewer-centered coordinates. To align their output with other methods, we transform all their results into object-centered coordinates by using camera pose. 
% We also abandon the iterative closest point algorithm (ICP) used in AtlasNet and TMNet because the ICP aligns the predicted and the ground-truth point sets, while we argue ground-truth shapes should be unavailable at test stage.

As illustrated by~\cref{table:comparison}, our approach using either of the two coordinates consistently outperforms the state-of-the-art methods on all metrics under both settings, and such advantage is especially pronounced on novel classes. Although Mesh R-CNN outperforms Pixel2Mesh quantitatively, we noticed that the reconstructed surface from Mesh R-CNN is not as smooth as Pixel2Mesh as proved by NC. Specifically, on novel classes, compared with the second best performer Mesh R-CNN, our \name\ significantly decreases CD from $8.84\times 10^{-3}$ to $6.69\times 10^{-3}$, NC from $7.50\times 10^{-1}$ to $5.80\times 10^{-1}$, and increases IoU from $74.10\times 10^{-2}$ to $81.50\times 10^{-2}$, F-score from $40.50\times 10^{-2}$ to $50.09\times 10^{-2}$. It is noted that on seen base classes, AtlasNet and TMNet perform better than the early work Pixel2Mesh, possibly due to their elaborated mesh deformation mechanism. However, these two methods need extensive training on a large amount of data, which means they heavily rely on priors, and are only able to process objects with similar structures, leading to their poor performance on unseen novel classes. We dive into this result and argue that: 1) elaborated mechanisms usually negatively affect generalization; 2) the global feature based reconstruction may function by recognition instead of generation, as pointed out in~\cite{whatdolearn}, leading to significant performance decline on data out of distribution, \ie, novel classes. The performance of SDFNet~\cite{sdfnet} also proves the weaknesses of global features, which is only slightly better than AtlasNet and TMNet on novel classes, although it is designed for generalized 3D reconstruction. In contrast, Pixel2Mesh, Mesh R-CNN and our \name\ take local features as the guidance of deformation to emphasize detailed geometry information. This strategy enforces the network to pay more attention to the shape structure. Also, some local structures are shared across objects, \eg, legs of sofas and tables, tires of cars and buses, and triggers of rifles and pistols, so the model could adapt well to unseen objects. Note that compared with Pixel2Mesh using only 2D local features, our \name\ uses both 2D and 3D local features, and the benefit of this strategy could be observed from \cref{tab:ablaton} in the ablation study. Moreover, our intermediate point cloud representation and  multi-view silhouette loss also guarantee the generalization and robustness on novel classes, so that our \name\ outperforms existing methods with a significant margin under both coordinates. \revise{It is noted that the performance of all methods drops on novel classes, as the task of reconstructing unseen classes is more challenging. Nonetheless, we would like to point out that, as shown in~\cref{table:comparison}, the performance of our model on \textbf{novel} classes is even better than the performance of some comparing methods on \textbf{base classes}. For example, our Chamfer Distance (CD) on \textbf{novel} classes ($6.69 \times 10^{-3}$) is even better than that of Pixel2Mesh, OccNet, and SDFNet on \textbf{base} classes; our F-score on \textbf{novel} classes ($50.09 \times 10^{-2}$) is even better than that of Pixel2Mesh, OccNet, Mesh R-CNN, and SDFNet on \textbf{base} classes; and our IoU score on \textbf{novel} classes ($81.50 \times 10^{-2}$) is even better than that of all other comparing methods on \textbf{base} classes.}

\cref{tab:pix3d} compares the model generalization from the synthetic ShapeNet dataset to the real Pix3D dataset. Although the large domain gap leads to serious performance decline, our approach still outperforms the state-of-the-art methods on all metrics, especially under CD our approach outperforms the second best methods Mesh R-CNN nearly $24\times 10^{-3}$. Such generalization ability verified the merits of our proposed contributions.

\revise{In~\cref{table:unsupervise,table:category_specific}, we compare unsupervised-learning methods (\eg, SDF-SRN~\cite{sdfsrn}, SS3D~\cite{ss3d}, SoftRas~\cite{SoftRasterizer}, TARS~\cite{tars3D}) with our proposed ~\name. For comparison with them, we apply the same model on DVR~\cite{DVR} ShapeNet dataset, and all images are rendered with cube normalization. The images of 13 categories (car, chair, monitor, plane, rifle, speaker, table, telephone, bench, cabinet, lamp, sofa, watercraft) are rendered by \textbf{DVR}~\cite{DVR} and the remaining 3 categories (bus, train, pistol) are rendered by ourselves with same parameters during rendering. Please note that, SDF-SRN~\cite{sdfsrn} and TARS~\cite{tars3D} were evaluated on category-specific reconstruction in their original papers. SS3D~\cite{ss3d} and SDF-SRN~\cite{sdfsrn} could be naturally extended to category-agnostic models for our cross-category reconstruction task, while TARS~\cite{tars3D} could not be used for cross-category reconstruction, as it relies on the jointly-trained category-specific ``canonical shape latent" which is unavailable for the unseen categories under our setting. For comparison, we conducted two types of experiments. First, we re-run SS3D~\cite{ss3d} and SDF-SRN~\cite{sdfsrn} for cross-category reconstruction under our setting, \ie, training on 8 base classes and evaluating on 8 base classes and 8 novel classes, respectively, as we did to our method. Specifically, for SDF-SRN~\cite{sdfsrn}, we transferred the category-specific SDF-SRN~\cite{sdfsrn} model into category-agnostic model by training on 8 bases classes, instead of only one class each model as in the original paper. For SS3D~\cite{ss3d}, we followed their training process (\ie, pre-training, self-training, distilling) but trained the model on the 8 base classes from ShapeNet for a fair comparison. Second, for TARS~\cite{tars3D}, since it could not be used for cross-category reconstruction, we therefore followed TARS~\cite{tars3D} to run our model for category-specific reconstruction on three object categories for comparison. As shown in~\cref{table:unsupervise}, our method demonstrates a clear superiority over SDF-SRN~\cite{sdfsrn} and SS3D~\cite{ss3d} in terms all evaluation metrics on both base classes and novel classes. Our method also outperforms TARS~\cite{tars3D}, SoftRas~\cite{SoftRasterizer}, and SDF-SRN~\cite{sdfsrn} in category-specific reconstruction as shown in~\cref{table:category_specific}, which is consistent across all three object categories. Note that our \name{} utilizes CD as supervision during training while these unsupervised learning methods only rely on differentiable rendering without 3D supervision.}

\begin{table*}[t]
\caption{\revise{Performance comparison on ShapeNet under Chamfer Distance (CD), Earth Mover Distance (EMD), Multi-view Silhouette IoU (IoU), Normal Consistency (NC), and F-score (F). Best results are bolded.}}
\label{table:unsupervise}
\centering
\resizebox{0.99\textwidth}{!}{
\revise{\begin{tabular}{c|ccccc|ccccc}
\hline
\multirow{2}{*}{Methods} & \multicolumn{5}{c|}{Base} & \multicolumn{5}{c}{Novel} \\
 \cline{2-11}
 & \makecell[c]{CD$\downarrow$\\$\times10^{-3}$} & \makecell[c]{EMD$\downarrow$\\$\times10^{-2}$} & \makecell[c]{IoU$\uparrow$\\$\times10^{-2}$} & \makecell[c]{NC$\downarrow$\\$\times10^{-1}$} & \makecell[c]{F-score$\uparrow$\\$\times10^{-2}$} & \makecell[c]{CD$\downarrow$\\$\times10^{-3}$} & \makecell[c]{EMD$\downarrow$\\$\times10^{-1}$} & \makecell[c]{IoU$\uparrow$\\$\times10^{-2}$} & \makecell[c]{NC$\downarrow$\\$\times10^{-2}$} & \makecell[c]{F-score$\uparrow$\\$\times10^{-2}$} \\
 \hline
SDF-SRN~\cite{sdfsrn} & 5.43 & 2.82 & 73.45 & 6.44 & 58.82 & 17.92 & 5.21 & 63.35 & 8.28 & 42.22\\
SS3D~\cite{ss3d} & 4.73 & 2.69 & 71.21 & 4.12 & 55.77 & 12.12 & 4.66 & 61.56 & 5.71 & 44.14\\
Ours (viewer) & \textbf{1.31}& \textbf{2.25}& \textbf{86.25}& \textbf{3.58}& \textbf{87.41}& \textbf{3.30}& \textbf{2.82}& \textbf{79.77}& \textbf{4.81}& \textbf{74.55} \\
 \hline
\end{tabular}
}}
\end{table*}
\begin{table*}[t]
\caption{\revise{Performance comparison of category-specific models on ShapeNet chair, plane, plane under Chamfer Distance (CD), Earth Mover Distance (EMD), Precision, Recall, and F-score. Best results are bolded.}}
\label{table:category_specific}
\centering
\resizebox{0.9\textwidth}{!}{
\revise{\begin{tabular}{c|c|ccc|c|ccc}
\hline
\multirow{2}{*}{Cat.} & \multirow{2}{*}{Method} & \multicolumn{3}{c|}{CD$\downarrow$} & EMD$\downarrow$ & Precision$\uparrow$ & Recall$\uparrow$ & F-score$\uparrow$ \\
 &  & acc. & cov. & overall &  & (\%) & (\%) & (\%) \\ \hline
\multirow{4}{*}{Car} & SoftRas~\cite{SoftRasterizer} & 0.372 & 0.302 & 0.337 & 0.723 & 93.04 & 96.62 & 94.80 \\
 & SDF-SRN~\cite{sdfsrn} & 0.141 & 0.144 & 0.142 & 0.452 & 99.76 & 99.84 & 99.80 \\
 & TARS~\cite{tars3D} & 0.141 & 0.140 & 0.140 & 0.446 & 99.70 & 99.81 & 99.75 \\
 & Ours (viewer) & \textbf{0.102} & \textbf{0.107} & \textbf{0.105} & \textbf{0.102} & \textbf{99.82} & \textbf{99.93} & \textbf{99.87} \\ \hline
\multirow{4}{*}{Chair} & SoftRas~\cite{SoftRasterizer} & 0.572 & 0.475 & 0.523 & 1.017 & 82.56 & 89.18 & 85.74 \\
 & SDF-SRN~\cite{sdfsrn} & 0.352 & 0.315 & 0.333 & 0.854 & 94.18 & 95.21 & 94.69 \\
 & TARS~\cite{tars3D} & 0.353 & 0.312 & 0.332 & 0.817 & 93.43 & 95.39 & 94.40 \\
 & Ours (viewer) & \textbf{0.185} & \textbf{0.157} & \textbf{0.171} & \textbf{0.285} & \textbf{98.83} & \textbf{99.16} & \textbf{98.98} \\ \hline
\multirow{4}{*}{Plane} & SoftRas~\cite{SoftRasterizer} & 0.215 & 0.207 & 0.211 & 0.588 & 98.74 & 98.42 & 98.58 \\
 & SDF-SRN~\cite{sdfsrn} & 0.193 & 0.154 & 0.173 & 0.576 & 98.55 & 99.11 & 98.83 \\
 & TARS~\cite{tars3D} & 0.194 & 0.152 & 0.173 & 0.533 & 98.79 & 99.34 & 99.06 \\
 & Ours (viewer) & \textbf{0.107} & \textbf{0.121} & \textbf{0.114} & \textbf{0.182} & \textbf{99.85} & \textbf{99.74} & \textbf{99.79}\\ \hline
\end{tabular}
}}
\end{table*}

\begin{figure*}[!t]
  \centering
    \includegraphics[width=0.95\textwidth]{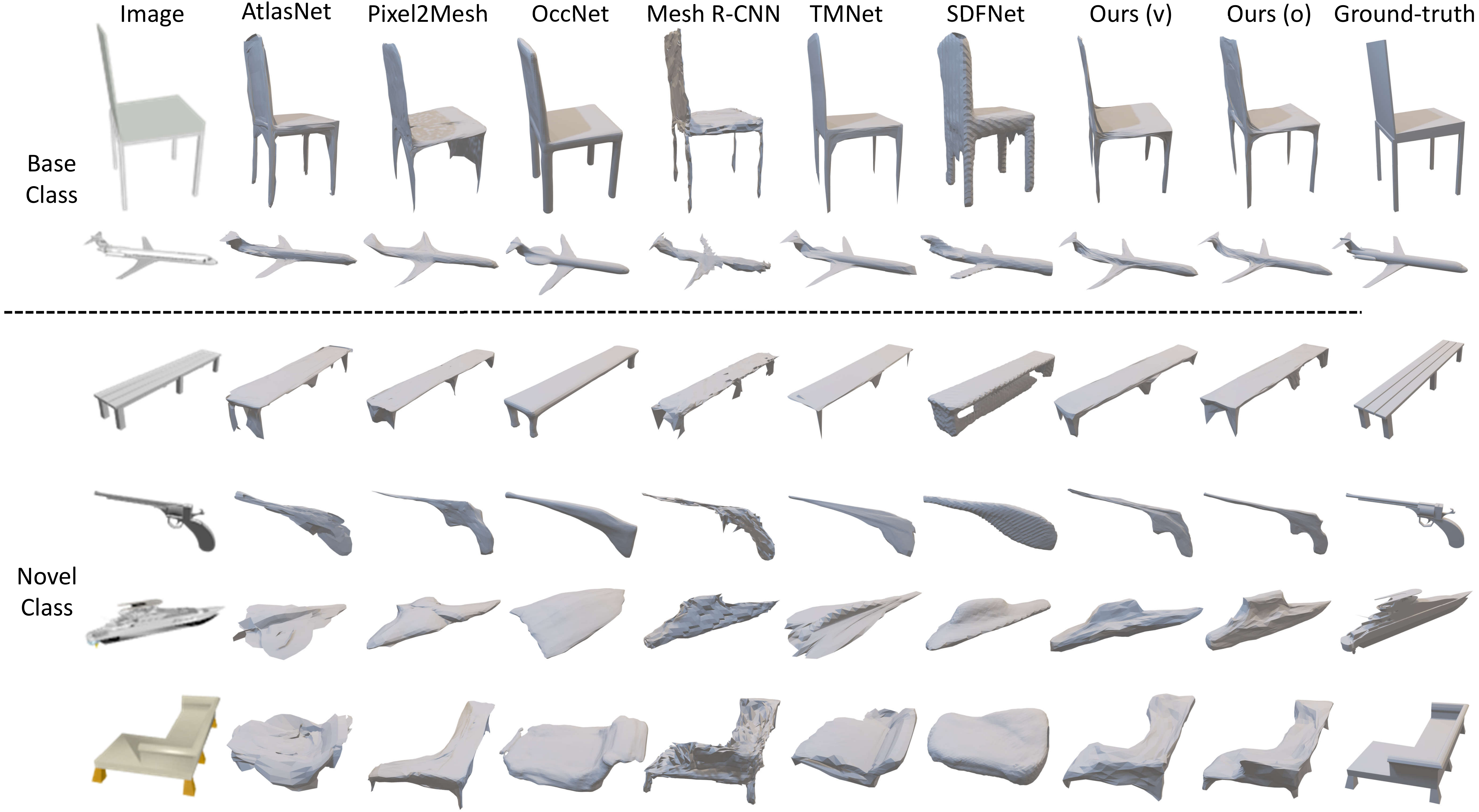}
    % \vspace{-2mm}
  \caption{Six visual examples (in rows) from ShapeNet~\cite{shapenet}. The 1st and 2nd rows are results on base classes, while the 3nd-6th rows are results on novel classes. The 1st and 9th columns show the input images and ground-truth shapes, and the 2nd-8th columns show the shapes reconstructed by Pixel2Mesh~\cite{pixel2mesh}, Mesh R-CNN~\cite{MeshRCNN}, AtlasNet~\cite{atlasnet}, TMNet~\cite{tmn}, SDFNet~\cite{sdfnet}, and Ours, respectively. 
  %Our \name\ outperforms previous methods significantly, especially on novel classes. Note that the global feature based AtlasNet~\cite{atlasnet} and TMNet~\cite{tmn} 
 %misidentify the watercraft as the plane at the 5nd row.
 }\label{fig:vis}
\end{figure*}

\begin{figure*}[!h]
%\vspace{-2mm}
  \centering
%   \fbox{\rule{0pt}{3.5in} \rule{.9\linewidth}{0pt}}
    \includegraphics[width=0.95\linewidth]{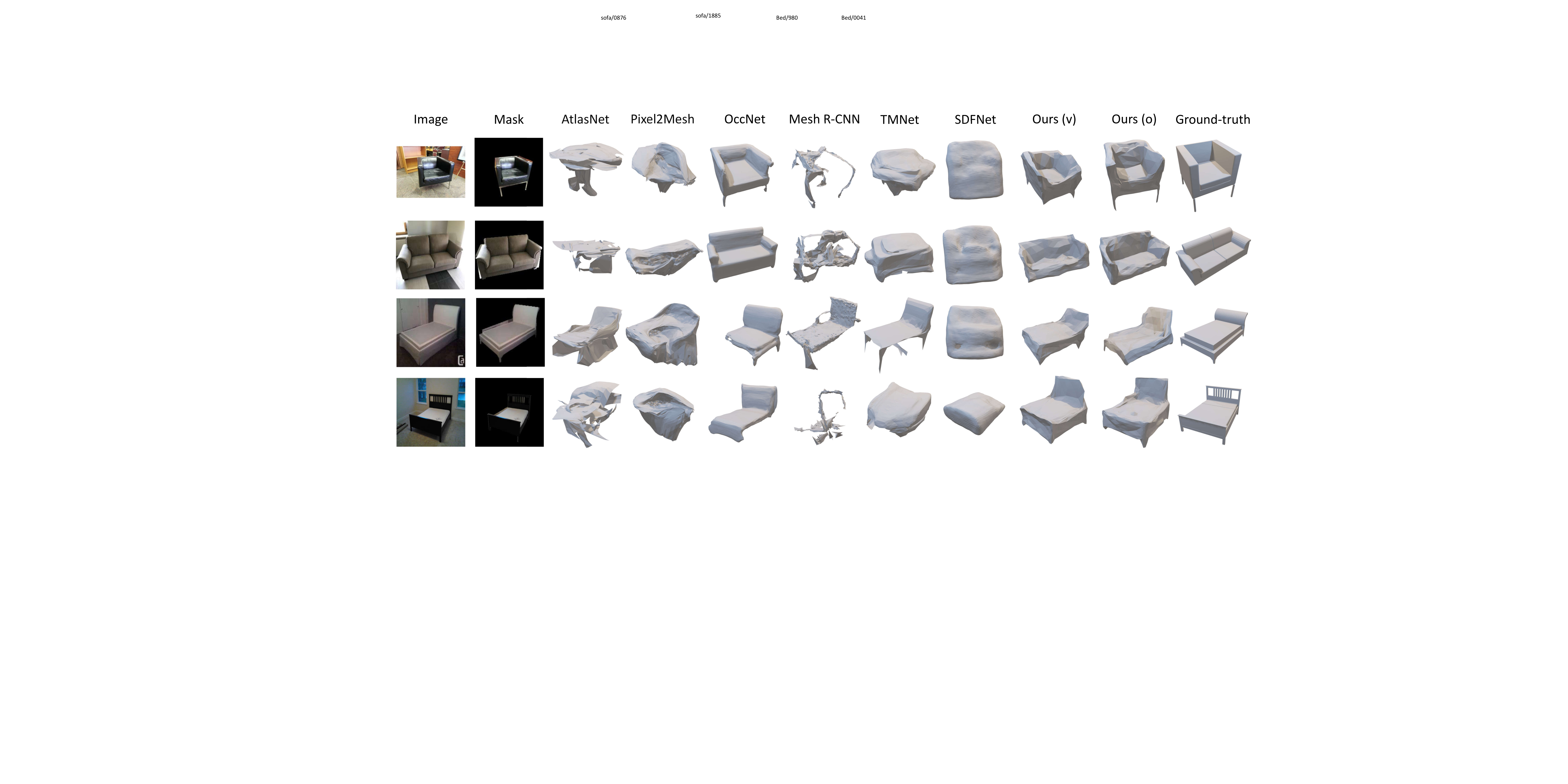}
    \vspace{-2mm}
  \caption{Visual reconstruction examples on real images from Pix3D~\cite{pix3d} dataset. The objects in the images are segmented by masks provided by Pix3D. All models are trained on ShapeNet and directly applied on Pix3D without any further training or refinement.}
  \label{fig:real}
\end{figure*}

\subsection{Qualitative Results.}
\cref{fig:vis} shows six visual examples of reconstruction results from \name\ and previous models, Pixel2Mesh~\cite{pixel2mesh}, Mesh R-CNN~\cite{MeshRCNN}, AtlasNet~\cite{atlasnet}, TMNet~\cite{tmn} and SDFNet~\cite{sdfnet}. For the base classes (the top two rows in \cref{fig:vis}), 
%Pixel2Mesh, AtlasNet, TMNet, SDFNet and our \name\ are all able to
all methods in comparison are able to generate accurate and smooth shapes. %and TMNet can yield better results on objects with complex topology, \eg, the chair. 
Our \name\ can achieve on-par performance with competitive works.
For novel classes (the 3rd to 6th rows in \cref{fig:vis}), if the target object is similar to the base classes (the 3rd and 4th rows), \eg, the bench and pistol similar to the chair and rifle, the outputs of AtlasNet and TMNet are not too bad but with visible reconstruction differences. If the input objects are very different from the base classes (the 5th and 6th rows), both AtlasNet and TMNet totally fail and even mis-identify the watercraft as the plane (the 5th row), implying they may function by recognition. Only Pixel2Mesh, Mesh R-CNN, SDFNet and our \name\ can generate reasonable results, while \name\ achieves the most faithful reconstruction. Note that although Mesh R-CNN outperforms Pixel2Mesh quantitatively, the former shows uneven surface and leads to worse visual results possibly due to its relatively weak mesh-based supervision.

Furthermore, to qualitatively evaluate the generalization ability on the real images, we test our \name\ and other methods on the Pix3D dataset~\cite{pix3d} by using the model trained on the ShapeNet~\cite{shapenet} without any further training or refinement. This is a challenging task as the images from Pix3D have remarkable differences with those from ShapeNet on both the style and the category. \cref{fig:real} shows the results from ours and the comparing methods, confirming the superiority of our method. \name\ is still able to reconstruct a variety of objects on real images faithfully, while most of the comparing methods fail in such a case.

\begin{table*}[!h]
  \centering
  \caption{Ablation study about the effects of different modules on ShapeNet. \textsl{Pixel2Mesh$^+$} is our baseline modified from Pixel2Mesh. All models reconstruct under object-centered coordinates.}
%   More details can be found in the supplement.}
  \label{tab:ablaton}
  \resizebox{0.9\textwidth}{!}{
  \begin{tabular}{@{}c|ccccc|ccccc@{}}
  \hline
    \multirow{2}{*}{Methods} & \multicolumn{5}{c|}{Base Classes} & \multicolumn{5}{c}{Novel Classes}\\ \cline{2-11}
%  & \multicolumn{1}{c|}{CD$\downarrow\times10^{-3}$} & \multicolumn{1}{|c|}{EMD$\downarrow\times10^{-2}$} & IoU$\uparrow\times10^{-2}$ & \multicolumn{1}{c|}{CD$\downarrow\times10^{-3}$} & \multicolumn{1}{|c|}{EMD$\downarrow\times10^{-2}$} & IoU$\uparrow\times10^{-2}$ \\ \hline
 & \multicolumn{1}{c|}{\makecell[c]{CD$\downarrow$\\$\times10^{-3}$}} & \multicolumn{1}{c|}{\makecell[c]{EMD$\downarrow$\\$\times10^{-2}$}} & \multicolumn{1}{c|}{\makecell[c]{IoU$\uparrow$\\$\times10^{-2}$}} & \multicolumn{1}{c|}{\makecell[c]{NC$\downarrow$\\$\times10^{-1}$}} & \multicolumn{1}{c|}{\makecell[c]{F$\uparrow$\\$\times10^{-2}$}} & \multicolumn{1}{c|}{\makecell[c]{CD$\downarrow$\\$\times10^{-3}$}} & \multicolumn{1}{c|}{\makecell[c]{EMD$\downarrow$\\$\times10^{-2}$}} & \multicolumn{1}{c|}{\makecell[c]{IoU$\uparrow$\\$\times10^{-2}$}} & \multicolumn{1}{c|}{\makecell[c]{NC$\downarrow$\\$\times10^{-1}$}} &
 \multicolumn{1}{c}{\makecell[c]{F$\uparrow$\\$\times10^{-2}$}}\\ \hline
 
  Pixel2Mesh$^+$ & \multicolumn{1}{c|}{4.47} & \multicolumn{1}{c|}{5.69} & \multicolumn{1}{c|}{82.69} & \multicolumn{1}{c|}{4.70} & \multicolumn{1}{c|}{56.24} & \multicolumn{1}{c|}{8.01} & \multicolumn{1}{c|}{5.96} & \multicolumn{1}{c|}{78.04} & \multicolumn{1}{c|}{6.14} & \multicolumn{1}{c}{45.87}\\ \hline
    w/o 2D local & \multicolumn{1}{c|}{6.58} & \multicolumn{1}{c|}{5.29} & \multicolumn{1}{c|}{79.68} & \multicolumn{1}{c|}{4.99} & \multicolumn{1}{c|}{50.97} & \multicolumn{1}{c|}{37.25} & \multicolumn{1}{c|}{10.71} & \multicolumn{1}{c|}{66.87} & \multicolumn{1}{c|}{7.24} & \multicolumn{1}{c}{35.48}\\
    w/o 3D local & \multicolumn{1}{c|}{4.09} & \multicolumn{1}{c|}{5.49} & \multicolumn{1}{c|}{85.44}  & \multicolumn{1}{c|}{4.55} & \multicolumn{1}{c|}{58.71} & \multicolumn{1}{c|}{7.07} & \multicolumn{1}{c|}{6.00} & \multicolumn{1}{c|}{80.80} & \multicolumn{1}{c|}{5.93} & \multicolumn{1}{c}{48.38} \\
    one-stage  & \multicolumn{1}{c|}{4.34} & \multicolumn{1}{c|}{5.58} & \multicolumn{1}{c|}{84.64}   & \multicolumn{1}{c|}{4.62} & \multicolumn{1}{c|}{57.23} & \multicolumn{1}{c|}{7.91} & \multicolumn{1}{c|}{6.24} & \multicolumn{1}{c|}{79.45} & \multicolumn{1}{c|}{6.03} & \multicolumn{1}{c}{46.67}\\
    w/o IoU loss & \multicolumn{1}{c|}{4.11} & \multicolumn{1}{c|}{5.59} & \multicolumn{1}{c|}{83.77} & \multicolumn{1}{c|}{4.58} & \multicolumn{1}{c|}{58.62} & \multicolumn{1}{c|}{6.94} & \multicolumn{1}{c|}{6.06} & \multicolumn{1}{c|}{80.06} & \multicolumn{1}{c|}{5.88} & \multicolumn{1}{c}{49.33} \\
    \hline
    Full Model  & \multicolumn{1}{c|}{\textbf{3.96}} & \multicolumn{1}{c|}{\textbf{5.36}} & \multicolumn{1}{c|}{\textbf{85.85}} & \multicolumn{1}{c|}{\textbf{4.51}} & \multicolumn{1}{c|}{\textbf{59.57}} & \multicolumn{1}{c|}{\textbf{6.69}} & \multicolumn{1}{c|}{\textbf{5.96}} & \multicolumn{1}{c|}{\textbf{81.50}} & \multicolumn{1}{c|}{\textbf{5.80}} & \multicolumn{1}{c}{\textbf{50.09}}\\
    \hline
  \end{tabular}}
\end{table*}

%
%-------------------------------------------------------------------------
\section{Discussion}
\subsection{Ablation Study}
\label{sec:ablation}

% \begin{figure}[t]
%   \centering
% %   \fbox{\rule{0pt}{1.5in} \rule{.9\linewidth}{0pt}}
%     \includegraphics[width=0.9\linewidth]{latex/curve.png}
%   \caption{Performance curve over weights of IoU loss $\lambda_{IoU}$. Each curve shows the mean CD, EMD and IoU for different weights. The baseline corresponds to $\lambda_{IoU}=0$, which means the IoU loss is removed.}
%   \label{fig:iou_weight}
% \end{figure}

% \begin{table}
%   \centering
%   \resizebox{\columnwidth}{!}{
%   \begin{tabular}{@{}lcccc@{}}
%     \toprule
%     Method &  {\bf CD$\downarrow\times10^3$} & {\bf EMD$\downarrow\times10^2$} & {\bf IoU$\uparrow\times10^2$} \\
%     \midrule
%     1-shot w/o IoU loss & 6.83 & 6.09 & 80.02 \\
%     1-shot 1-stage  & - & - & - \\
%     \midrule
%     global 2d + global 3d  & - & - & - \\
%     local 2d + global 3d  & - & - & - \\
%     global 2d + local 3d  & - & - & - \\
%     \midrule
%     Ours  & - & - & - \\
%     \bottomrule
%   \end{tabular}
%   }
%   \caption{1-shot Ablation Study.}
%   \label{tab:1shot}
% \end{table}

We conduct ablative studies to single out the improvement from each of our claimed contributions. Firstly, to demonstrate the local features on generalization, we compare the performance of our model by using either local features or global features. Secondly, we re-design a one-stage version of our proposed network to demonstrate the benefits from the intermediate point cloud representation on 3D mesh reconstruction. Thirdly, we compare our model performance with or without the proposed multi-view silhouette loss. Finally, we provide a vanilla baseline ``Pixel2Mesh$^+$" to verify the improvement collectively. This is a modified Pixel2Mesh model, 
%For fair comparison, we also provide a modified Pixel2Mesh$^+$ which directly outputs shapes under object-centered coordinates as our local-feature-based object-centered baseline in~\cref{tab:ablaton}, 
where the graph convolution used in Pixel2Mesh is replaced by MLPs and better training recipe is applied, consistent with our \name. Compared with our ``Full model", ``Pixel2Mesh$^+$" only employs the 2D local features like Pixel2Mesh, without any of other proposed components.

\noindent\textbf{Local vs Global.} To use global features, we pool the feature maps or feature groups from the 2D or 3D encoders into feature vectors, and concatenate them with vertex coordinates for the following deformation, same as AtlasNet and TMNet. The results show that global features yield obviously poor performance on novel classes, which is comparable to AtlasNet and TMNet in \cref{tab:ablaton}, verifying the necessity of sampling local features. The dramatic performance decline of \name\ without 2D local features is due to the unreasonable intermediate representation, which invalidates the 3D feature sampling at the second stage and further degrades the local feature model into the global feature model.
% \textcolor{red}{Especially, the local 2D features are more important compared with 3D local features, as the performance decrease more dramatically.}
%global features are too coarse to distinguish details, while local features can preserve geometry information that are shared cross categories and prevent recognition. Especially, the local 2D features are more important compared with 3D local features, as the performance decrease more dramatically.

\noindent\textbf{Two-stage vs one-stage.}~~To verify the two-stage architecture, we develop a one-stage variant of our model without using the intermediate point cloud generation. In particular, we sample the local features from the image plane according to the vertex coordinates of the mesh template,
%and the camera intrinsic matrix and pose. 
 and concatenate the local features with the vertex coordinates to get the per-vertex feature representation. Lacking the intermediate point clouds, we cannot sample 3D features. By comparing ``one-stage" and ``Full Model" in \cref{tab:ablaton}, the performance of the one-stage model significantly decreases on all criteria, verifying the advantages of the proposed two-stage reconstruction for model generalization.

\noindent\textbf{Multi-view Silhouette Loss.}~~We investigate the effect of our proposed multi-view silhouette loss, by comparing our full framework with a variant without using the IoU loss. By comparing ``w/o IoU loss" with ``Full Model" , and ``Pixel2Mesh$^+$" with ``one-stage" in \cref{tab:ablaton}, the employment of IoU loss for training not only yields improvement on the IoU metric for evaluation, but also reduces CD and EMD, on both the base and the novel classes. 
%Moreover, by adjusting the loss weight $\lambda_{IoU}$. As can be seen in \cref{fig:iou_weight}, . This experiment demonstrates the value of our novel loss as a general strategy to strengthen 3D mesh reconstruction.

\noindent\textbf{Comparison with the baseline ``Pixel2Mesh$^+$"}.~~As mentioned, ``Pixel2Mesh$^+$" is our vanilla baseline, similar to the original Pixel2Mesh~\cite{pixel2mesh}. With the Multi-view Silhouette Loss, 3D local features and intermediate point cloud generation, our ``Full Model" outperforms ``Pixel2Mesh$^+$" under all evaluation protocols, further demonstrating the merits of our collective design.

\subsection{Sensitivity to Object Scales.}
%  Regarding the scale of 3D objects, following previous works ~\cite{atlasnet,tmn}, we normalize 3D objects by a unit sphere before evaluation, so that their scale will not influence the evaluation metrics. Regarding the scale of objects in 2D images, we investigate its influence by rendering the same object at varied depths and the results in \cref{tab:depth} show novel classes are more sensitive to depth variation than base classes, while our method remains reasonably robust to the scale of 2D objects in the images.
To investigate the effect of object scales to reconstruction quality, we re-render the same objects with same parameters except depth. Specifically, we scale the depth with 1.01, 1.05, and 1.10, respectively. We evaluate our model on the three test sets directly without any further training to verify its robustness to object scales. The quantitative comparison in ~\cref{tab:depth} shows that novel classes are more sensitive to depth variation than base classes, while our method remains reasonably robust to the scale of 2D objects in the images.

\begin{figure*}[!t]
  \centering
%   \fbox{\rule{0pt}{0.9\linewidth} \rule{0.9\linewidth}{0pt}}
  \includegraphics[width=0.95\textwidth]{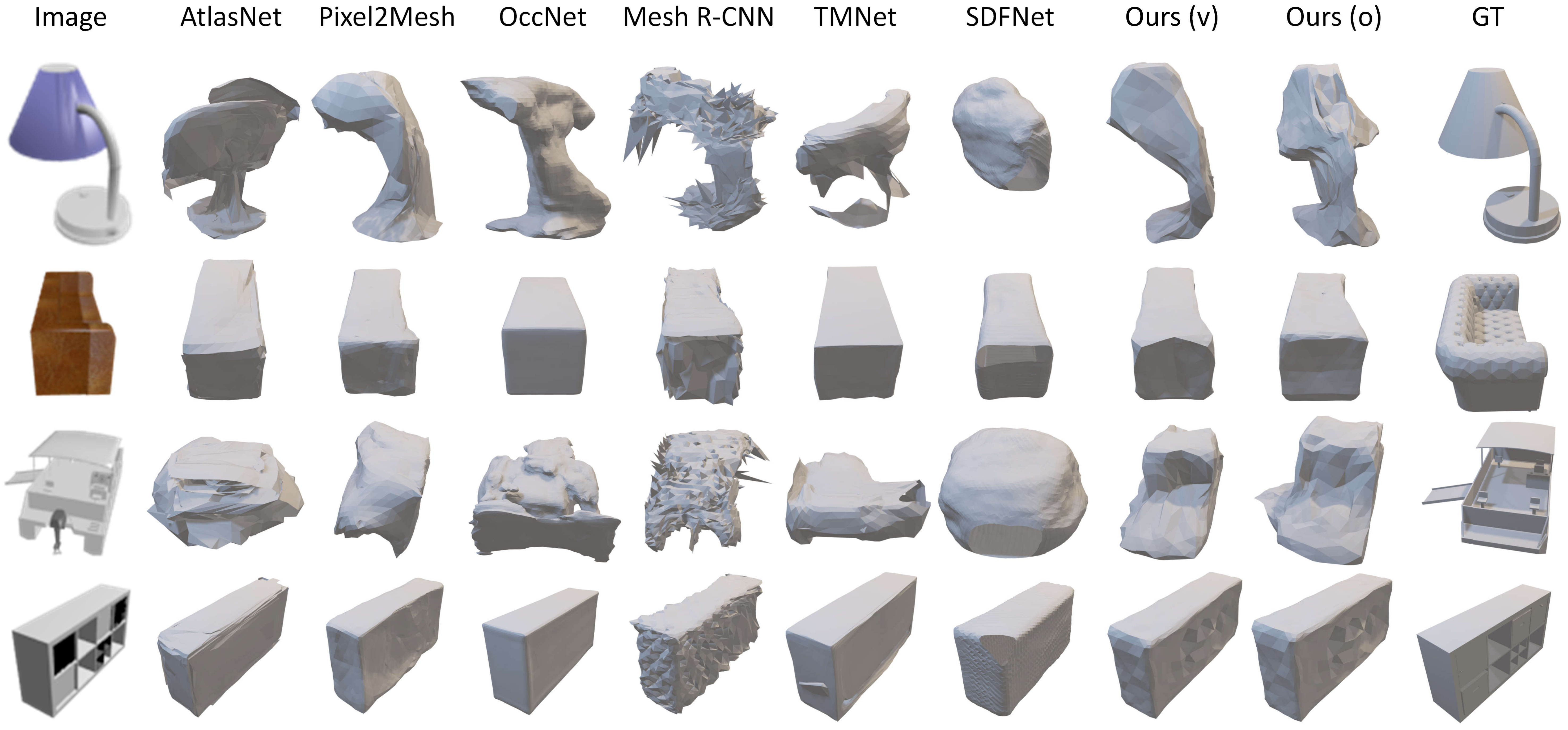}
   \caption{Four failure examples from ShapeNet~\cite{shapenet}. Both existing methods and our \name\ occasionally fail to reconstruct thin lamp pole (Row 1), blur boundaries (Row 2), and complex topology (Row 3 and Row 4).}
   \label{fig:fail}
\end{figure*}

\begin{table}[t]
    \centering
    \caption{Sensitivty to Object Scales on ShapeNet~\cite{shapenet}.}
    \label{tab:depth}
    \resizebox{0.45\textwidth}{!}{
    \begin{tabular}{@{}c|c|c|c|c|c|c@{}}
    \hline
        % \multirow{2}{*}{Methods} & \multicolumn{2}{3||}{Base} & \multicolumn{2}{c}{Novel} \\ \cline{2-5} 
         \multicolumn{2}{c|}{Methods} & CD$\downarrow$ & EMD$\downarrow$ & IoU$\uparrow$  & NC$\downarrow$ & F-score$\uparrow$ \\
        \hline
         \multirow{3}{*}{Base} & original & 3.96 & 5.36 & \textbf{85.85} & \textbf{4.51} & 59.57 \\
         & depth +1\% & \textbf{3.95} & 5.36 & 85.85 & 4.51 & \textbf{59.72} \\
         & depth +5\% & 3.99 & 5.36 & 85.83 & 4.51 & 59.59 \\
         & depth +10\% & 4.02 & \textbf{5.35} & 85.75 & 4.51 & 59.52 \\
         \hline
         \multirow{3}{*}{Novel} & original & \textbf{6.69} & 5.96 & \textbf{81.50} & 5.80 & \textbf{50.09} \\
         & depth +1\% & 6.75 & \textbf{5.92} & 81.46 & 5.80 & 50.03 \\
         & depth +5\% & 6.81 & 5.95 & 81.46 & \textbf{5.79} & 50.00 \\
         & depth +10\% & 7.04 & 6.03 & 81.30 & 5.91 & 49.74 \\
         \hline
    \end{tabular}
    }
\end{table}

% \begin{figure}[t]
%   \centering
% %   \fbox{\rule{0pt}{0.45\textwidth} \rule{0.9\linewidth}{0pt}}
%   \includegraphics[width=0.49\textwidth]{depth.pdf}
%   \caption{The reconstruction results of \name\ to different depth (original, depth+$1\%$, depth+$5\%$, depth+$10\%$).}
%   \label{fig:depth}
% \end{figure}

\subsection{\revise{Sensitivity to Image View.}}

\begin{figure}[h]
\centering
\includegraphics[width=0.95\linewidth]{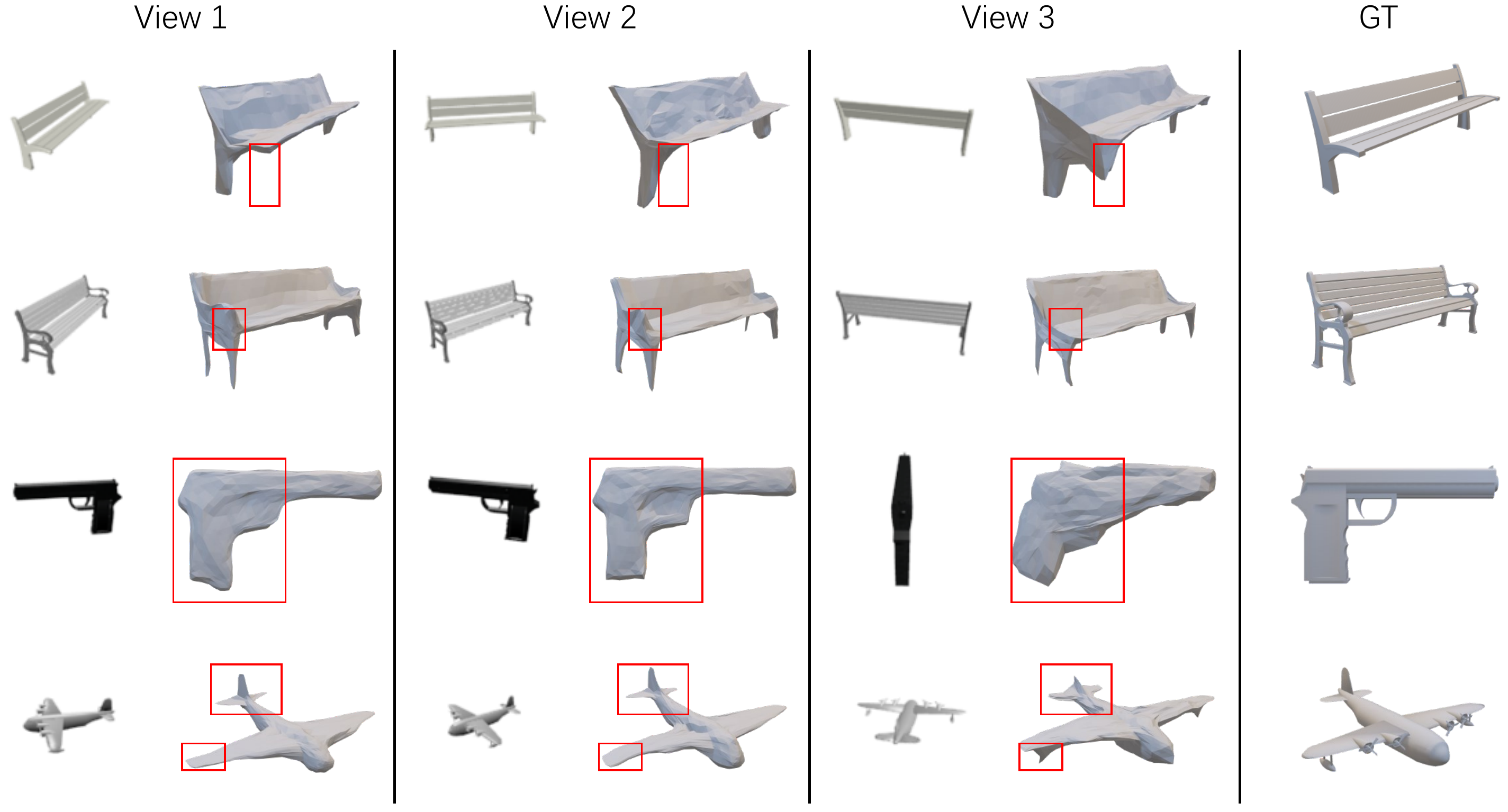}
% \rule{16cm}{8cm}
\caption{\revise{Visual reconstruction comparison from different viewpoints.}}
\label{fig:viewpoint}
\end{figure}

\revise{To assess the model's sensitivity to viewpoint variations, we compare the reconstruction results of the same object under different viewpoints. Specifically, for each object we render four additional images from random viewpoints, forming four additional datasets, and conduct five experiments (Exp 1$\sim$5) to evaluate our trained model on these five datasets (the original and the four newly generated ones), respectively. The quantitative results are shown in~\cref{table:view_sensitive}, while four visual examples are given~\cref{fig:viewpoint}.  From~\cref{fig:viewpoint}, we can see the change of viewpoints does affect the reconstruction results as some views (\eg, View 3 where most of object details could not be observed) capture much less information than others (\eg, View 1). The results illustrate that the larger the visible portion of the object and the more the discriminate information captured in the input image, the better the reconstruction results. Even for some unfavorable cases, (\eg, pistol under View 3 in~\cref{fig:viewpoint}), our model still yields reasonable reconstruction results possibly because the missing details of the object could be somewhat learned from other objects with a better viewpoint in the same category. On the other hand, from the quantitative results in~\cref{table:view_sensitive}, we can see the variance of the five experiment results is insignificant across all the evaluation metrics, as either ideal or extremely bad viewpoints are rarely sampled compared with other viewpoints in between.}

\begin{table*}[t]
\caption{\revise{Performance sensitivity to viewpoints on ShapeNet under Chamfer Distance (CD), Earth Mover Distance (EMD), Multi-view Silhouette IoU (IoU), Normal Consistency (NC), and F-score.}}
\label{table:view_sensitive}
\centering
\resizebox{0.99\textwidth}{!}{
\revise{\begin{tabular}{c|ccccc|ccccc}
\hline
\multirow{2}{*}{Method} & \multicolumn{5}{c|}{Base} & \multicolumn{5}{c}{Novel} \\
 \cline{2-11}
& \makecell[c]{CD$\downarrow$\\$\times10^{-3}$} & \makecell[c]{EMD$\downarrow$\\$\times10^{-2}$} & \makecell[c]{IoU$\uparrow$\\$\times10^{-2}$} & \makecell[c]{NC$\downarrow$\\$\times10^{-1}$} & \makecell[c]{F-score$\uparrow$\\$\times10^{-2}$} & \makecell[c]{CD$\downarrow$\\$\times10^{-3}$} & \makecell[c]{EMD$\downarrow$\\$\times10^{-1}$} & \makecell[c]{IoU$\uparrow$\\$\times10^{-2}$} & \makecell[c]{NC$\downarrow$\\$\times10^{-2}$} & \makecell[c]{F-score$\uparrow$\\$\times10^{-2}$} \\
 \hline
Exp 1 & 3.96 & 5.36 & 85.85 & 4.51 & 59.57 & 6.69 & 5.96 & 81.50 & 5.80 & 50.09 \\
Exp 2 & 3.86 & 5.36 & 86.06 & 4.49 & 60.37 & 6.50 & 5.94 & 81.69 & 5.76 & 50.25  \\
Exp 3 & 4.13 & 5.40 & 85.79 & 4.51 & 59.79 & 6.70 & 5.95 & 81.72 & 5.78 & 50.25 \\
Exp 4 & 4.25 & 5.51 & 85.56 & 4.55 & 59.41 & 6.89 & 6.04 & 81.53 & 5.82 & 50.31 \\
Exp 5 & 3.96 & 5.41 & 85.97 & 4.48 & 59.85 & 6.71 & 5.90 & 81.71 & 5.78 & 50.38 \\
\hline
% mean  & 4.03 & 5.41 & 85.84 & 4.51 & 59.80 & 6.70 & 5.96 & 81.63 & 5.79 & 50.26 \\
% sigma & 0.15 & 0.06 & 0.19 & 0.03 & 0.36 & 0.14 & 0.05 & 0.10 & 0.02 & 0.11 \\
Mean$\pm$STD  & $4.03\pm0.15$ & $5.41\pm0.06$ & $85.84\pm0.19$ & $4.51\pm0.03$ & $59.80\pm0.36$ & $6.70\pm0.14$ & $5.96\pm0.05$ & $81.63\pm0.10$ &$5.79\pm0.02$ & $50.26\pm0.11$ \\
 \hline
\end{tabular}
}}
\end{table*}

\subsection{\revise{Joint-training vs Off-the-shelf}}
\revise{We adopted the off-the-shelf networks for point generation~\cite{3DAttriFlow} and surface reconstruction~\cite{atlasnet} to emphasize the importance of the joint-training strategy. The results are given in~\cref{table:off_the_shelf}. These two released models were trained on 13 object classes in ShapeNet. Therefore, some ``novel" classes (\ie, bench, cabinet, lamp, sofa, watercraft) in our setting are not completely novel to these two models, \ie, the setting favors the off-the-shelf models. As shown in~\cref{table:off_the_shelf}, our model outperforms the two-stage pipeline using the off-the-shelf models~\cite{3DAttriFlow,atlasnet} in a large margin in terms of all metrics. Both the base classes and the novel classes could benefit from our joint-training and special design for single-view reconstruction.}

\begin{table*}[t]
\caption{\revise{Performance comparison on ShapeNet under Chamfer Distance (CD), Earth Mover Distance (EMD), Multi-view Silhouette IoU (IoU), Normal Consistency (NC), and F-score. Best results are bolded.}}
\label{table:off_the_shelf}
\centering
\resizebox{0.99\textwidth}{!}{
\revise{\begin{tabular}{c|ccccc|ccccc}
\hline
\multirow{2}{*}{Methods} & \multicolumn{5}{c|}{Base} & \multicolumn{5}{c}{Novel} \\
 \cline{2-11}
& \makecell[c]{CD$\downarrow$\\$\times10^{-3}$} & \makecell[c]{EMD$\downarrow$\\$\times10^{-2}$} & \makecell[c]{IoU$\uparrow$\\$\times10^{-2}$} & \makecell[c]{NC$\downarrow$\\$\times10^{-1}$} & \makecell[c]{F-score$\uparrow$\\$\times10^{-2}$} & \makecell[c]{CD$\downarrow$\\$\times10^{-3}$} & \makecell[c]{EMD$\downarrow$\\$\times10^{-1}$} & \makecell[c]{IoU$\uparrow$\\$\times10^{-2}$} & \makecell[c]{NC$\downarrow$\\$\times10^{-2}$} & \makecell[c]{F-score$\uparrow$\\$\times10^{-2}$} \\
 \hline
\makecell{Off-the-shelf~\cite{3DAttriFlow,atlasnet}} & 9.68 & 7.86 & 73.04 & 4.98 & 43.03 & 18.77 & 9.78 & 69.53 & 6.52 & 38.37 \\
Ours (v)  & 4.18 & 5.54 & 83.91 & 4.67 & 57.23 & 6.71 & 6.16 & 79.85 & 5.96 & 47.61 \\
Ours (o) & \bf{3.96} & \bf{5.36} & \bf{85.85} & \bf{4.51} & \bf{59.57} & \bf{6.69} & \bf{5.96} & \bf{81.50} & \bf{5.80} & \bf{50.09}\\
 \hline
\end{tabular}
}}
\end{table*}

\subsection{Limitation} Examples of failure cases are provided in \cref{fig:fail}. As shown, existing methods and our \name{} may occasionally fail on thin structures (\eg, lamp pole), blur boundaries (\eg, sofa armrest), and complex topology (\eg, boat, cabinet), which are common challenges for the community. First, the failure on thin structures is mainly caused by two reasons: 1) point number imbalance in training the loss function, and 2) deformation difficulty. These can be mitigated by raising the importance of points from thin structures to force the model to pay more attention to thin structures during deformation. Second, the blur boundaries are introduced by the datasets. It's an ill-posed problem for single-view 3D reconstruction and can only be mitigated by learned priors. Third, by using the sphere template, our \name\ could keep the surface closed, but this also restricts the deformation of objects with holes (\eg, cabinet) and complex topology (\eg, boat). This issue could be handled by a post-processing step to treat holes specially in the literature~\cite{tmn}. We would like to point out that such a restriction could also be possibly relaxed by using multiple spheres to form the template meshes. This could not only generate more complex structures but also keep the surface closed, and will be explored in our future study. Note that existing methods that could avoid the topology problem theoretically (\ie, AtlasNet~\cite{atlasnet}, OccNet~\cite{occnet}, Mesh R-CNN~\cite{MeshRCNN}, TMNet~\cite{tmn}, SDFNet~\cite{sdfnet}) also fail on these cases. 

\revise{Like many existing methods for 3D single-view reconstruction, our model does not explicitly handle occlusion. This is because reconstructing a 3D object from a single 2D image is an ill-posed problem, making it difficult to determine the visibility of a projected point. Since all 3D points that are projected along the same ray will be projected onto the same point on the 2D image plane, these points obtain the same 2D features without differentiating whether or not they are occluded. It would be an interesting research direction to develop an explicit algorithm for processing occluded points in future work.}

%------------------------------------------------------------------------

\section{Conclusion}
In this paper, we put forward a learning framework, \name, to solve \revise{single-view 3D mesh reconstruction on seen and unseen categories}. We propose three strategies to improve the model generalization ability on novel classes and prevent overfitting, namely, learning intermediate point cloud representation, employing local features, and introducing multi-view silhouette loss for model regularization. Our model demonstrates a promising capacity for cross-category object reconstruction and generalizes to unseen object classes well.

\section{Grant Acknowledgement}
This research is partly supported by Australian Research Council (ARC) DP200103223 and the Ministry of Education, Singapore, under its Academic Research Fund Tier 2 (MOE-T2EP20220-0007) and Tier 1 (RG14/22). 


\section{References Section}
\bibliographystyle{IEEEtran}
\bibliography{egbib}

% \end{thebibliography}

\newpage
\section{Biography Section}
 

\vspace{-320pt}
\begin{IEEEbiography}[{\includegraphics[width=1in,height=1.25in,clip,keepaspectratio]{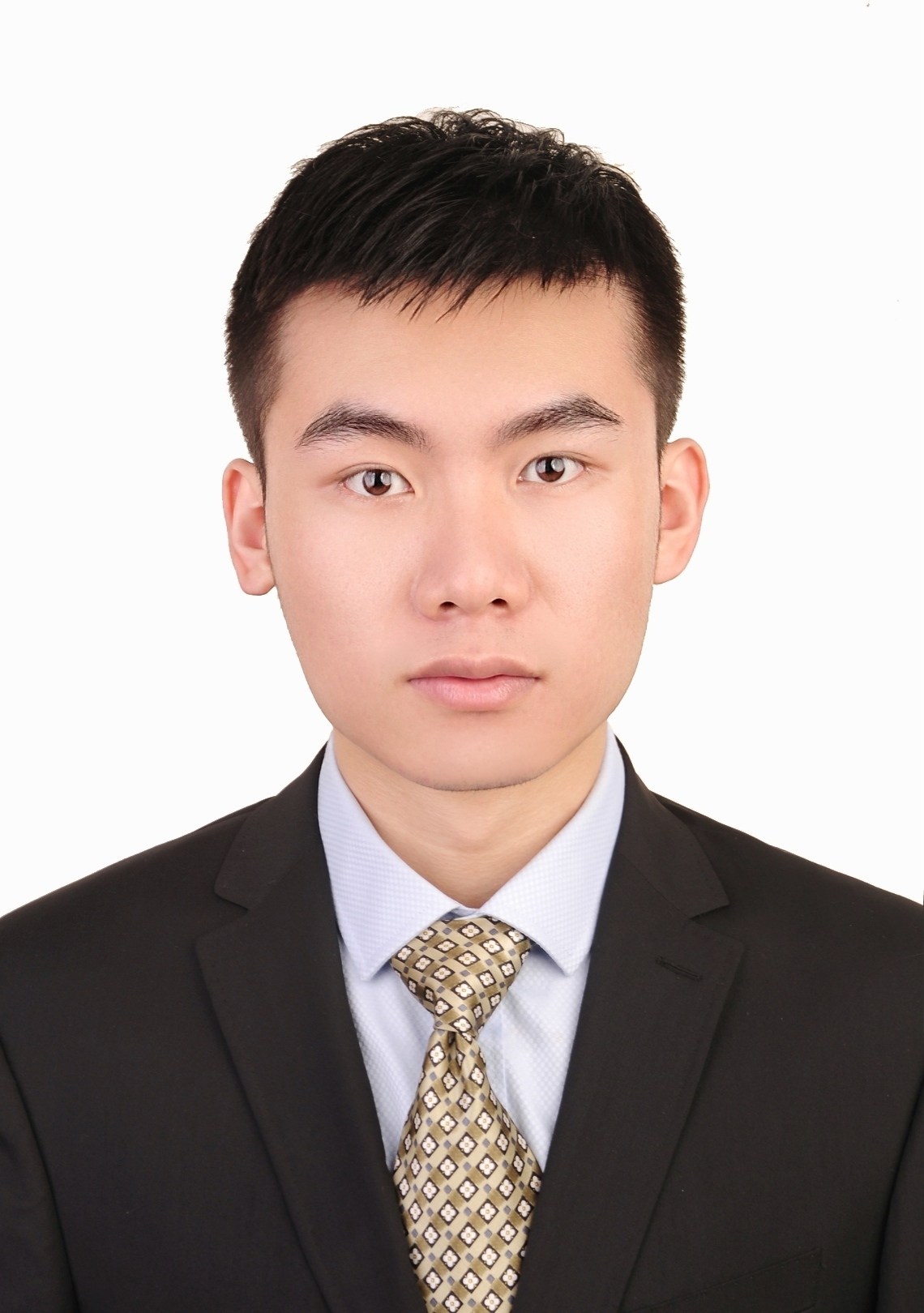}}]{Xianghui Yang} is a Ph.D. candidate in the School of Electrical \& Information Engineering, The University of Sydney, where he works at the USYD-Vision Lab under the supervision of Prof. Luping Zhou, Prof. Guosheng Lin and Prof. Wanli Ouyang. Before that, he received B.Sc. degree in Physiscs from the School of Physics, Nanjing University in 2019. His research interests include 3D reconstruction and surface reconstruction.
% \vspace{-150pt}
\end{IEEEbiography}

\vspace{-320pt}
\begin{IEEEbiography}[{\includegraphics[width=1in,height=1.25in,clip,keepaspectratio]{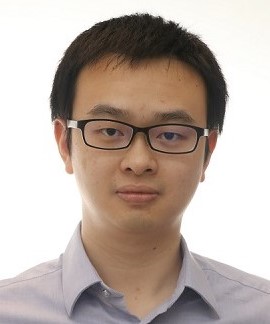}}]{Guosheng Lin} is an Assistant Professor at the School of Computer Science and Engineering, Nanyang Technological University, Singapore. He received his PhD degree from The University of Adelaide in 2014. His research interests are in computer vision and machine learning.
\end{IEEEbiography}

\vspace{-320pt}
\begin{IEEEbiography}[{\includegraphics[width=1in,height=1.25in,clip,keepaspectratio]{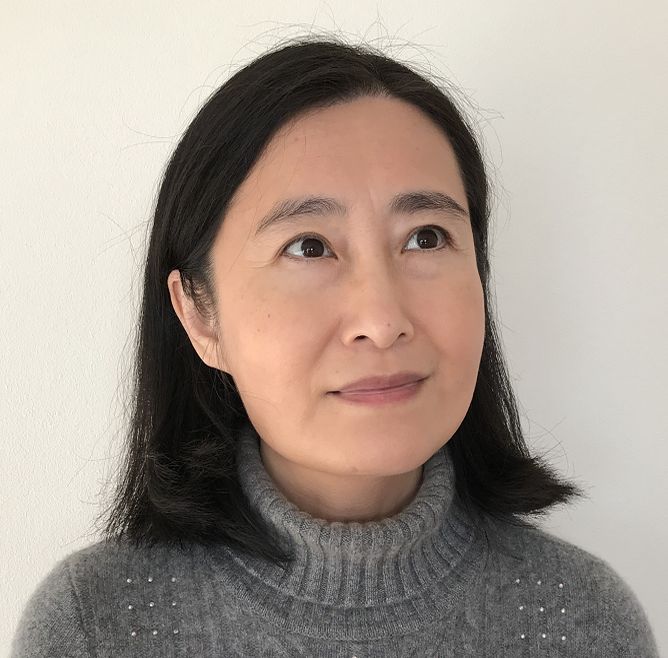}}]{Luping Zhou} is now an Associate Professor in School of Electrical and Information Engineering at the University of Sydney, Australia. She received her PhD from Australian National University, and she was a recipient of Australian Research Council DECRA award (Discovery Early Career Researcher Award) in 2015. Her research interests include medical computer vision, machine learning, and medical image analysis. She is a senior member of IEEE.
\end{IEEEbiography}



% \vfill

\end{document}